\documentclass[runningheads]{llncs}
\usepackage{comment}
\usepackage{amsmath,amssymb,bbm} % define this before the line numbering.
\usepackage{color}
\newcommand{\myfirstpara}[1]{\noindent {\bf #1.}}
\newcommand{\mypara}[1]{\vspace{0.5em} \noindent {\bf #1.}}
\usepackage{epsfig}
\usepackage{graphicx}
\usepackage{mathtools}
\usepackage{breqn}
\usepackage{multirow,enumitem,float,placeins,chngcntr}
\usepackage[hyperfootnotes=false]{hyperref}
\usepackage{booktabs} % {color,mathtools,comment}
\usepackage{wrapfig}
\usepackage{algorithm,algpseudocode}

\newcommand\blfootnote[1]{%
	\begingroup
	\renewcommand\thefootnote{}\footnote{#1}%
	\addtocounter{footnote}{-1}%
	\endgroup
}

\newcommand{\bx}{\mbox{$\mathbf{x} $}}
\newcommand{\bI}{\mbox{$\mathbf{I} $}}

\newcommand{\calI}{\mbox{$\mathcal{I} $}}

\newcommand{\calR}{\mbox{$\mathcal{R} $}}
\newcommand{\btheta}{\mbox{\boldmath${\theta} $}}
\newcommand{\bP}{\mbox{\boldmath${P}$}}

\def\H{{\mathbb{H}}}
\def\1{{\mathbbm{1}}}
\DeclarePairedDelimiterX{\infdivx}[2]{(}{)}{%
	#1\;\delimsize\|\;#2%
}
\newcommand{\kl}{\text{KL}\infdivx}

\DeclarePairedDelimiter{\norm}{\lVert}{\rVert}
\DeclarePairedDelimiter{\card}{\lvert}{\rvert}

\DeclareMathOperator*{\argmax}{arg\,max}

\begin{document}
% \linenumbers

\pagestyle{headings}
\mainmatter
\def\ECCVSubNumber{2510}  % Insert your submission number here

\title{Contextual Diversity for Active Learning} % Replace with your title

\begin{comment}
\titlerunning{ECCV-20 submission ID \ECCVSubNumber} 
\authorrunning{ECCV-20 submission ID \ECCVSubNumber} 
\author{Anonymous ECCV submission}
\institute{Paper ID \ECCVSubNumber}
\end{comment}
%******************

%\begin{comment}
\titlerunning{Contextual Diversity for Active Learning}
\author{Sharat Agarwal$ ^\ast $\inst{1} \and
Himanshu Arora$ ^\ast $$^\dagger $\inst{2} \and
Saket Anand\inst{1} \and 
Chetan Arora\inst{3}}

\authorrunning{S. Agarwal et al.}

\institute{IIIT-Delhi, India, ~~\email{\{sharata,anands\}@iiitd.ac.in} \and
Flixstock Inc., ~~ \email{himanshu@flixstock.com}\\
\and
Indian Institute of Technology Delhi, India, ~~ \email{chetan@cse.iitd.ac.in}}
%\end{comment}
%******************

\maketitle
\blfootnote{$ ^\ast $Equal contribution.} 
\blfootnote{$ ^\dagger $Work done while the author was at IIIT-Delhi.}\setcounter{footnote}{0}

\begin{abstract}
	
Requirement of large annotated datasets restrict the use of deep convolutional neural networks (CNNs) for many practical applications. The problem can be mitigated by using active learning (AL) techniques which, under a given annotation budget, allow to select a subset of data that yields maximum accuracy upon fine tuning. State of the art AL approaches typically rely on measures of visual diversity or prediction uncertainty, which are unable to effectively capture the variations in spatial context. On the other hand, modern CNN architectures make heavy use of spatial context for achieving highly accurate predictions. Since the context is difficult to evaluate in the absence of ground-truth labels, we introduce the notion of contextual diversity that captures the confusion associated with spatially co-occurring classes. Contextual Diversity (CD) hinges on a crucial observation that the probability vector predicted by a CNN for a region of interest typically contains information from a larger receptive field. Exploiting this observation, we use the proposed CD measure within two AL frameworks: (1) a core-set based strategy and (2) a reinforcement learning based policy, for active frame selection. Our extensive empirical evaluation establish state of the art results for active learning on benchmark datasets of Semantic Segmentation, Object Detection and Image classification. Our ablation studies show clear advantages of using contextual diversity for active learning. The source code and additional results will soon be available at \url{https://github.com/sharat29ag/CDAL}. 

\end{abstract}

\begin{figure}[t]
	\centering
	\includegraphics[scale=0.375]{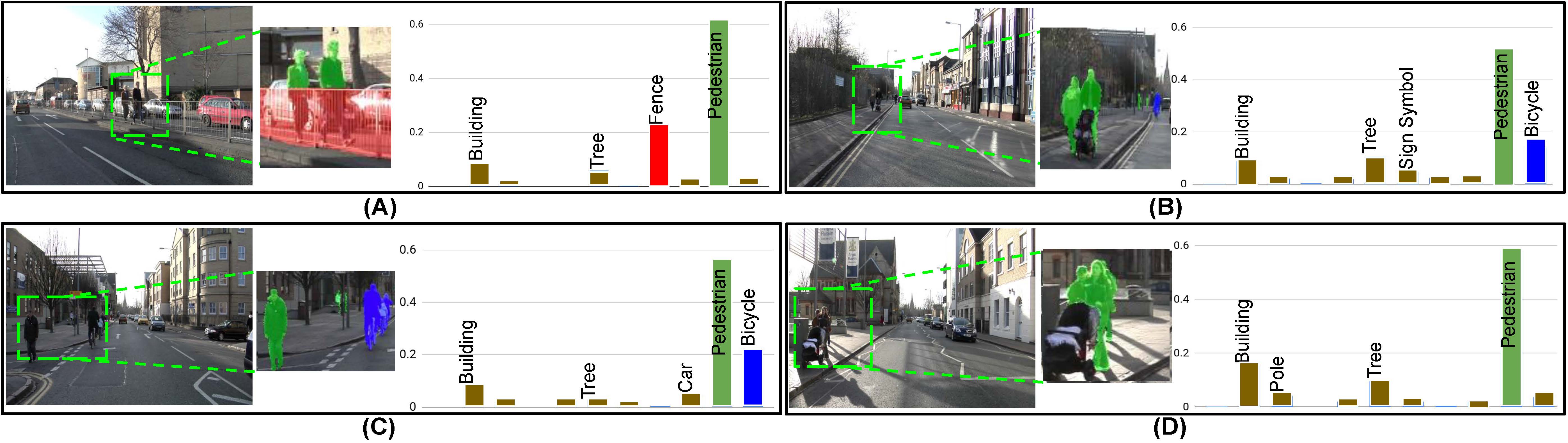}
	\caption{ Illustration showing 4 frames from Camvid. Each subfigure shows the full RGB image, region of interest with ground truth overlaid, and the average probability for the `pedestrian' class with bars color coded by class. We observe that the confusion reflected by the average probability vector corresponding to a class in a frame is also influenced by the object's background. Notice the confusion of pedestrian class with fence in (A) and with bicycle in (C), each of which appear in the neighborhood of a pedestrian instance. We propose a novel \emph{contextual diversity} based measure that exploits the above structure in probability vectors to help select images containing objects in diverse backgrounds. Including this set of images for training helps improving accuracy of CNN-based classifiers, which rely on the local spatial neighborhoods for prediction. For the above example our contextual diversity based selection picks $ \{(A), (C), (D)\} $ as opposed to the set $ \{(B), (C), (D)\} $ picked by a maximum entropy based strategy (best viewed in color).} 
	\label{fig:inc_class_context_div}
\end{figure}

\section{Introduction}
\label{sec:intro}

Deep convolutional neural networks (CNNs) have acheived state of the art (SOTA) performance on various computer vision tasks. One of the key driving factors for this success has been the effort gone in preparing large amounts of labeled training data. As CNNs become more popular, they are applied to diverse tasks from disparate domains, each of which may incur annotation costs that are task as well as domain specific. For instance, the annotation effort in image classification is substantially lower than that of object detection or semantic segmentation in images or videos. Similarly, annotations of RGB images may be cheaper than MRI/CT images or Thermal IR images, which may require annotators with specialized training. 

The core idea of Active Learning (AL) is to leverage the current knowledge of a machine learning model to select most \emph{informative} samples for labeling, which would be more beneficial to model improvement compared to a randomly chosen data point \cite{settles2012active}. With the effectiveness of deep learning (DL) based models in recent years, AL strategies have been investigated for these models as well. Here, it has been shown that DL models trained with a fraction of available training samples selected by active learning can achieve nearly the same performance as when trained with all available data \cite{sener2018active,Yoo_2019_CVPR,VAAL_ICCV2019}. Since DL models are expensive to train, AL strategies for DL typically operate in a batch selection setting, where a set of images are selected and annotated followed by retraining or fine-tuning of the model using the selected set. 

Traditional AL techniques \cite{lewis1994heterogeneous,lewis1994sequential,settles2008analysis,luo2013latent,joshi2009multi} have mostly been based on \emph{uncertainty} and have exploited the ambiguity in the predicted output of a model. As most measures of uncertainty employed are based on predictions of individual samples, such approaches often result in highly correlated selections in the batch AL setting. Consequently, more recent AL techniques attempt to reduce this correlation by following a strategy based on the \emph{diversity} and \emph{representativeness} of the selected samples \cite{Wang_CSVT2017,yang2017suggestive,Kuo_MICCAI2018}. Existing approaches that leverage these cues are still insufficient in adequately capturing the spatial and semantic context within an image and across the selected set. Uncertainty, typically measured through entropy, 
%can capture spatial context when computed locally, but is unable
is also unable 
to capture the class(es) responsible for the resulting uncertainty. On the other hand, visual diversity and representativeness are able to capture the semantic context across image samples, but are typically measured using global cues in a feature space that do not preserve information about the spatial location or relative placement of the image's constituent objects. 
% the sense that visual diversity is a global cue and is typically measured in a feature space that does not capture the spatial location or relative placements of its constituent objects. Though, uncertainty, typically measured through entropy, could be location-specific, it does not capture the classes it arises from. 

Spatial context is an important aspect of modern CNNs, which are able to learn discriminative semantic representations due to their large receptive fields. There is sufficient evidence that points to the brittleness of CNNs as object locations, or the spatial context, in an image are perturbed \cite{Rosenfeld_ArXivAug2018_ElephantRoom}. In other words, a CNN based classifier's misclassification is not simply attributed to the objects from the true class, but also to other classes that may appear in the object's spatial neighborhood. This crucial observation also points to an important gap in the AL literature, where existing measures are unable to capture uncertainty arising from the diversity in spatial and semantic context in an image. Such a measure would help select a training set that is diverse enough to cover a \emph{variety of object classes} and their \emph{spatial co-occurrence} and thus improve generalization of CNNs. The objective of this paper is to achieve this goal by designing a novel measure for active learning which helps select frames having objects in diverse contexts and background. Figure \ref{fig:inc_class_context_div} describes an illustrative comparison of some of the samples selected by our approach with the entropy based one.

% One way to improve generalization of CNNs is by selectively taking a training set that is diverse enough to cover a \emph{variety of object classes} and their \emph{spatial co-occurrence} in the training images. The objective of this paper is to achieve the same by designing a novel reward for the active learning which helps select frames having objects in diverse contexts and background. Figure \ref{fig:inc_class_context_div} compares some of the samples selected by our approach with the uncertainty based one.

In this paper, we introduce the notion of contextual diversity, which permits us to unify the model prediction uncertainty with the diversity among samples based upon spatial and semantic context in the data. % We propose a new information-theoretic distance measure based on the idea, which allows us to use it as a surrogate distance measure within the core-set based approach \cite{sener2018active}. We also show its utility as a reward function in a reinforcement learning (RL) framework and show further improvements in the performance of active frame selection. Through various ablative experiments, we demonstrate that it can complement existing measures of visual diversity and representation. 
We summarize our contributions below: 
\begin{itemize}
\item We introduce a novel information-theoretic distance measure, Contextual Diversity (CD), to capture the diversity in spatial and semantic context of various object categories in a dataset.
\item We demonstrate that using CD with core-set based active learning \cite{sener2018active} almost always beats the state of the art across three visual recognition tasks: semantic segmentation, object detection and image classification. We show an improvement of 1.1, 1.1, and 1.2 units on the three tasks, over the state of the art performance achieving 57.2, 73.3, and 80.9 respectively.
\item Using CD as a reward function in an RL framework further improves the AL performance and achieves an improvement of 2.7, 2.1, and 2.3 units on the respective visual recognition tasks over state of the art (57.2, 73.3, and 80.9 respectively).
\item Through a series of ablation experiments, we show that CD complements existing cues like visual diversity.
\end{itemize}  

\section{Related Work}\label{sec:rel_work}

Active learning techniques can be broadly categorized into the following categories. Query by committee methods operate on consensus by several models \cite{beluch2018power,gal2017deep}. However, these approaches in general are too computationally expensive to be used with deep neural networks and big datasets. Diversity-based approaches identify a subset of a dataset that is sufficiently representative of the entire dataset. Most approaches in this category leverage techniques like clustering \cite{nguyen2004active}, matrix partitioning \cite{guo2010active}, or linkage based similarity\cite{bilgic2009link}. Uncertainty based approaches exploit the ambiguity in the predicted output of a model. Some representative techniques in this class include \cite{lewis1994heterogeneous,lewis1994sequential,settles2008analysis,luo2013latent,joshi2009multi}. Some approaches attempt to combine both uncertainty and diversity cues for active sample selection. Some notable works in this category include \cite{Li_CVPR2013,Wang_CSVT2017,yang2017suggestive,Kuo_MICCAI2018}. Recently, generative models have also been used to synthesize informative samples for Active Learning   \cite{zhu2017generative,mayer2018adversarial,mahapatra2018efficient}. In the following, we give a detailed review of three recent state of the art AL approaches applied to vision related tasks. We compare with these methods later in the experiment sections over the three visual recognition tasks. 

\mypara{Core-Set} 
Sener and Savarese \cite{sener2018active} have modeled active learning as a \emph{core-set} selection problem in the feature space learned by convolutional neural networks (CNNs). Here, the core-set is defined as a selected subset of points such that the union of $ \mathbb{R}^n$-balls of radius $ \delta $ around these points contain \emph{all} the remaining unlabeled points. The main advantage of the method is in its theoretical guarantees, which claim that the difference between the loss averaged over all the samples and that averaged over the selected subset does not depend on the \emph{number of samples} in the selected subset, but only on the radius $\delta$. Following this result, Sener and Savarese used approximation algorithms to solve a facility location problem using a Euclidean distance measure in the feature space. However, as was noted by \cite{VAAL_ICCV2019}, reliance on Euclidean distance in a high-dimensional feature space is ineffective. Our proposed contextual diversity measure relies on KL divergence, which is known to be an effective surrogate for distance in the probability space \cite{Dabak_RicePhDThesis_1992}. Due to distance like properties of our measure, the proposed approach, named \emph{contextual diversity based active learning using core-sets} (CDAL-CS), respects the theoretical guarantees of core-set, yet does not suffer from curse of dimensionality. 

\mypara{Learning Loss} 
Yoo and Kweon \cite{Yoo_2019_CVPR} have proposed a novel measure of uncertainty by learning to predict the loss value of a data sample. They sampled data based on the ranking obtained on the basis of predicted loss value. However, it is not clear if the sample yielding the largest loss, is also the one that leads to most performance gain. The samples with the largest loss, could potentially be outliers or label noise, and including them in the training set may be misleading to the network. The other disadvantage of the technique is that, there is no obvious way to choose the diverse samples based upon the predicted loss values. 

\mypara{Variational Adversarial Active Learning (VAAL)}	
Sinha et al. \cite{VAAL_ICCV2019} have proposed to use a VAE to map both the labeled and unlabeled data into a latent space, followed by a discriminator to distinguish between the two based upon their latent space representation. The sample selection is simply based on the output probability of the discriminator. Similar to \cite{Yoo_2019_CVPR}, there seem to be no obvious way to choose diverse samples in their technique based on the discriminator score only. Further, there is no guarantee that the representation learnt by their VAE is closer to the one used by the actual model for the task. Therefore, the most informative frame for the discriminator need not be the same for the target model as well. Nonetheless, in the empirical analysis, VAAL demonstrates state of the art performance among the other active learning techniques for image classification and semantic segmentation tasks. 

\mypara{Reinforcement Learning for Active Learning}
Recently, there has been an increasing interest in application of RL based methods to active learning. RALF \cite{RALF_CVPR2012} takes a meta-learning approach where the policy determines a weighted combination of pre-defined uncertainty and diversity measure, which is then used for sample selection. Both \cite{Woodward_NIPSwDRL2016} and \cite{DRAL_ICCV2019} train the RL agents using ground truth based rewards for one-shot learning and person re-identification separately. This requires their method to have a large, annotated training set to learn the policy, and therefore is hard to generalize to more annotation heavy tasks like object detection and semantic segmentation. In \cite{Konyushkova_2018_CVPR}, an RL framework minimizes time taken for object-level annotation by choosing between bounding box verification and drawing tasks. Fang et al. \cite{Fang_EMNLP2017} design a special state space representation to capture uncertainty and diversity for active learning for text data. This design makes it harder to generalize their model to other tasks. Contrary to most of these approaches, our RL based formulation, CDAL-RL, takes a task specific state representation and uses the contextual diversity based reward that combines uncertainty and diversity in an unsupervised manner.

%\mypara{Active Learning for Semantic Segmentation} 
%
\begin{figure}[t]
	\begin{center}
		\includegraphics[width=0.7\linewidth]{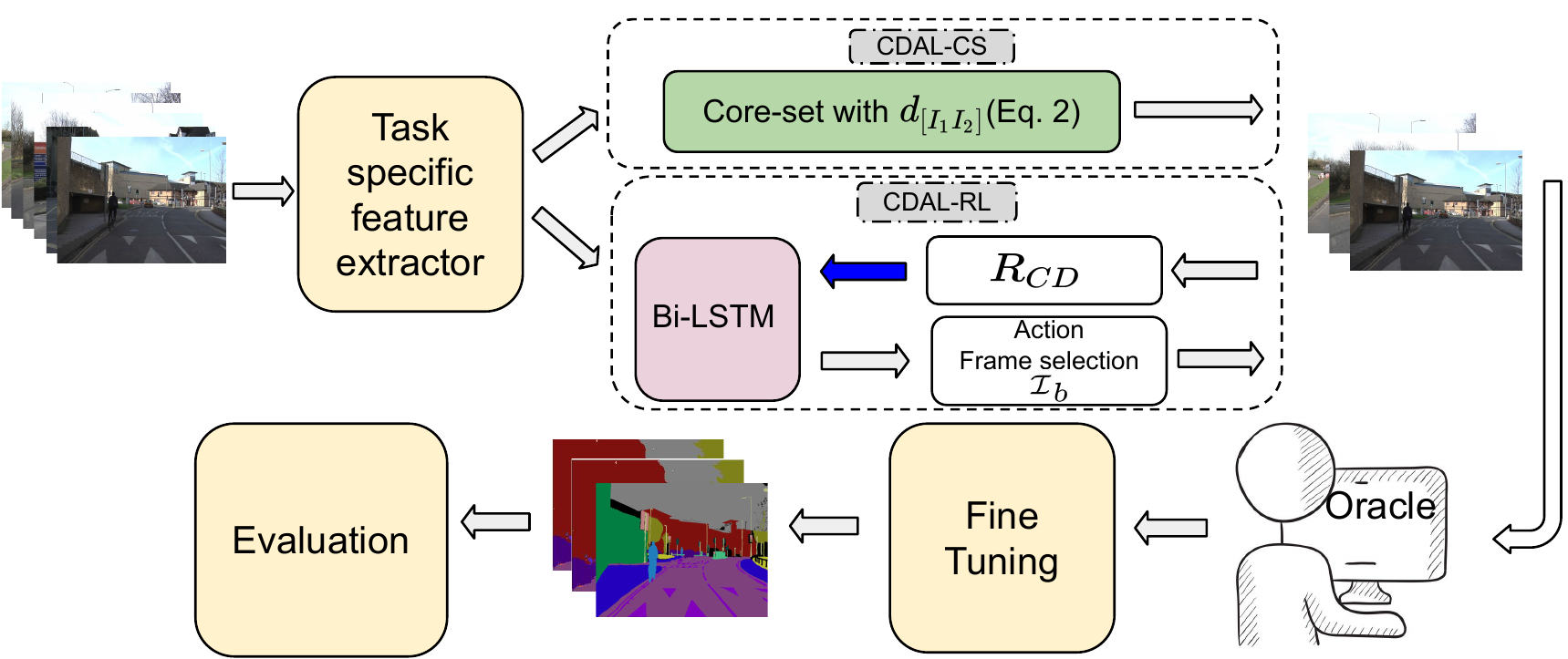}
	\end{center}
	\caption{The architecture for the proposed frame selection technique. Two of the strengths of our technique are its unsupervised nature and its generalizability to variety of tasks. The frame selection can be performed in either way by CDAL-CS or CDAL-RL modules. Based on the visual task, a pre-trained model can be readily integrated. The top scoring frames are selected to be annotated and are used to fine tune the model to be evaluated over the main task.  %The reward function comprises terms of contextual diversity, semantic representation and visual representation which improve the generalization despite selecting fewer frames. The unsupervised nature stems from the use of this reward function in an RL framework. Based on the visual task, a pre-trained model can be readily integrated with an LSTM based policy network, which in turn produces a score for each frame. The top scoring frames are selected to be annotated and are used to fine tune the model. Note that the RL module is \emph{completely unsupervised}, and does not need any annotation. The fine-tuned model is then evaluated over the main task.
	}
	\label{fig:arch}
\end{figure}

\section{Active Frame Selection}\label{sec:meth}

One of the popular approaches in semi-supervised and self-supervised learning is to use \emph{pseudo-labels}, which are labels as predicted by the current model \cite{Caron_Clustering_ECCV2018,lee2013pseudo,Wang_CSVT2017}. However, directly using pseudo-labels in training, without appropriately accounting for the uncertainty in the predictions could lead to overfitting and confirmation bias \cite{arazo2019pseudolabeling}. Nonetheless, the class probability vector predicted by a model contains useful information about the model's discriminative ability. In this section, we present the proposed \emph{contextual diversity} (CD), an information-theoretic measure that forms the basis for Contextual Diversity based Active Learning (CDAL). At the heart of CD is our quantification of the model's predictive uncertainty defined as a mixture of softmax posterior probabilities of pseudo-labeled samples. This mixture distribution effectively captures the spatial and semantic context over a set of images. We then derive the CD measure, which allows us to select diverse and uncertain samples from the unlabeled pool for annotation and finally suggest two strategies for active frame selection. First (CDAL-CS), inspired by the core-set \cite{sener2018active} approach and the second (CDAL-RL) using a reinforcement learning framework. An overview of our approach to Active Learning is illustrated in Fig. \ref{fig:arch}.

%One of the popular approaches in semi-supervised and self-supervised learning is to use \emph{pseudo-labels}, which are labels as predicted by the current model \cite{Caron_Clustering_ECCV2018,lee2013pseudo,Wang_CSVT2017}. However, directly using pseudo-labels in training, without appropriately accounting for the uncertainty in the predictions could lead to overfitting and confirmation bias \cite{arazo2019pseudolabeling}. In this section, we quantify the predictive uncertainty of the model by defining a mixture of softmax posterior probabilities of pseudo-labeled samples. We then derive the \emph{contextual diversity} measure, which allows us to select diverse and uncertain samples from the unlabeled pool for annotation, which forms the basis for Contextual Diversity based Active Learning (CDAL). Finally, we suggest two strategies for active frame selection: first (CDAL-CS), inspired by the core-set \cite{sener2018active} approach and the second (CDAL-RL) using a reinforcement learning framework. Our approach to Active Learning is illustrated in Fig. \ref{fig:arch}.

\subsection{Contextual Diversity}\label{sec:CDAFSN}

Deep CNNs have large receptive fields to capture sufficient spatial context for learning discriminative semantic features, however, it also leads to feature interference making the output predictions more ambiguous \cite{Rosenfeld_ArXivAug2018_ElephantRoom}. This spatial pooling of features adds to confusion between classes, especially when a model is not fully trained and has noisy feature representations. We quantify this ambiguity by defining the \emph{class-specific confusion}. 

Let $ C = \{1,\ldots,n_C \} $ be the set of classes to be predicted by a Deep CNN based model. Given a region $ r $ within an input image $ \bI $, let $ \bP_r(\widehat{y} \mid \bI ; \btheta)$ be the \emph{softmax probability vector} as predicted by the model $ \btheta $. For convenience of notation, we will use $ \bP_r $ instead of $ \bP_r(\widehat{y}|\bI;\btheta) $ as the subscript $ r $ implies the conditioning on its constituent image $ \bI $ and the model $ \btheta $ is fixed in one step of sample selection. These regions could be pixels, bounding boxes or the entire image itself depending on the task at hand. The pseudo-label for the region $ r \subseteq \bI $ is defined as $\widehat{y}_r = \argmax_{j\in C} \bP_r[j]$, where the notation $ \bP_r[j] $ denotes the $ j^{th} $ element of the vector. We emphasize that this abstraction of regions is important as it permits us to define overlapping regions within an image and yet have different predictions, thereby catering to tasks like object detection. Let $ \calI=\cup_{\tiny c\in C}\calI^c$ be the total pool of \emph{unlabeled} images, where $ \calI^c $ is the set of images, each of which have at least one region classified by the model into class $ c $. Further, let $ \calR_{\tiny\bI}^c $ be the set of regions within image $ \bI\in\calI^c $ that are assigned a pseudo-label $ c $. The collection of \emph{all} the regions that the model believes belong to class $ c $ is contained within the set $ \calR_{\tiny\calI}^c = \cup_{\tiny \bI\in\calI^c}\calR^c_{\tiny\bI} $. We assume that for a sufficiently large unlabeled pool $ \calI $, there will be a non-empty set $ \calR_{\tiny\calI}^c$. For a given model $ \btheta $ over the unlabeled pool $ \calI $, we now define the class-specific confusion for class $ c $ by the following mixture distribution $ \bP_{\tiny \calI}^c $ 
%We now define the class-specific confusion of a given model $ \btheta $, for a class $ c $, over the unlabeled pool $ \calI $, using the the following mixture distribution $ \bP_{\tiny \calI}^c $ 
\begin{equation}
\label{eqn:class_cont_conf}
\bP_{\tiny \calI}^c = \frac{1}{|\calI^c|} \sum_{\bI\in \calI^c} \left[ \frac{\sum_{\tiny r \in \calR_{\tiny \bI}^c} w_{r} \bP_r(\widehat{y} \mid \bI; \btheta)}{\sum_{\tiny r\in \calR_{\tiny \bI}^c}{w_r}} \right]
\end{equation} 
with $ w_r \geq 0$ as the mixing weights. While the weights could take any non-negative values, we are interested in capturing the predictive uncertainty of the model. Therefore, we choose the weights to be the Shannon entropy of $w_r= \H(\bP_r)=-\sum_{j\in C} \bP_r[j] \log_2 \bP_r[j] +\epsilon$, where $ \epsilon > 0$ is a small constant and avoid any numerical instabilities.
If the model were perfect, $ \bP_{\tiny\calI}^c $ would be a one-hot encoded vector\footnote{We ignore the unlikely event where the predictions are perfectly consistent over the large unlabeled pool $ \calI $, yet different from the \emph{true} label.}, but for an insufficiently trained model $ \bP_{\tiny\calI}^c $ will have a higher entropy indicating the confusion between class $ c $ and all other classes ($ c^\prime\in C, c^\prime\neq c $). We use $ \bP^c_{\tiny \bI} $ to denote the mixture computed from a single image $ \bI\in \calI^c $.

%Intuitively, the probability vector $ \bP_{\tiny\calI}^c $ captures the overall confusion of the model between class $ c $ and others over the pool $ \calI $. 
As discussed in Sec. \ref{sec:intro}, in CNN based classifiers, this uncertainty stems from spatial and semantic context in the images. For instance, in the semantic segmentation task shown in Fig. \ref{fig:inc_class_context_div}, the model may predict many pixels as of class `pedestrian' ($ c=\text{pedestrian} $) with the highest probability, yet it would have a sufficiently high probability of another class like `fence' or `bicycle'. In such a case, $ \bP_{\tiny\calI}^{c}[j] $ will have high values at $ j = \{\text{fence, bicycle}\} $, reflecting the chance of confusion between these classes across the unlabeled pool $ \calI $. As the predictive ability of the model increases, we expect the probability mass to get more concentrated at $j\!=\!c$ and therefore reduce the overall entropy of $ \bP_{\tiny\calI}^c $. It is easy to see that the total Shannon's entropy $ h_{\tiny\calI}=\sum_{c\in C} \H(\bP_{\tiny\calI}^c) $ reduces with the cross-entropy loss. 
 
% {\color{red} We show that $ H_{\tiny\calI}=\sum_{c\in C}H(P_{\tiny\calI}^c) $, i.e., the entropy of $ P_{\tiny\calI}^c $ minimized if and only if the cross-entropy loss is minimized \textbf{under mild assumptions}\footnote{what are these assumptions?}.}

%Therefore, with $ H_{\tiny \calI} = \sum_{(c\in C)}H(P^c) $ as the surrogate loss, given budget $ b $, our goal for active frame selection is to form $ \calI_b $, a set of $ b $ frames, which when incorporated into the training set, leads to maximum reduction in $ H_{\tiny \calI}$. 
% Pairwise KL-divergence has been used for estimating entropy and mutual information for mixture distributions \cite{Kolchinsky_2017}.
Annotating an image and using it to train a model would help resolve the confusion constituent in that image. Based on this intuition, we argue that the annotation effort for a new image is justified only if its inclusion increases the informativeness of the selected subset, i.e., when an image captures a \emph{different kind} of confusion than the rest of the subset. Therefore, for a given pair of images $ \bI_1$ and $ \bI_2 $, we quantify the disparity between their constituent class-specific confusion by the \emph{pairwise contextual diversity} defined using a symmetric KL-divergence as 
% We make use of a symmetrized version of KL-divergence to capture the disparity between class-specific confusion of two images $ \bI_1 $ and $ \bI_2 $ quantified by the pairwise contextual diversity 
\begin{equation}
\label{eqn:KL_div}
d_{\tiny [\bI_{1},\bI_{2}]} = \sum_{c\in C} \1^c(\bI_{1},\bI_{2}) \left(0.5*\kl{\bP_{\tiny\bI_1}^c}{\bP_{\tiny\bI_2}^c} + 0.5*\kl{\bP_{\tiny\bI_2}^c}{\bP_{\tiny \bI_1}^c}\right).
\end{equation}
In Eq. (\ref{eqn:KL_div}), $ \kl{\cdot}{\cdot} $ denotes the KL-divergence between the two mixture distributions. We use the indicator variable denoted by $ \1^c(\cdot) $ that takes a value of one only if both $ \bI_1,\bI_2\in\calI^c $, otherwise zero. This variable ensures that the disparity in class-specific confusion is considered only when both images have at least one region pseudo-labeled as class $ c $, i.e., when both images have a somewhat reliable measure of confusion w.r.t. class $ c $. This confusion disparity accumulated over all classes is the pairwise contextual diversity measure between two images. Given that the KL-divergence captures a distance between two distributions, $ d_{\tiny [\bI_{1},\bI_{2}]} $ can be used as a distance measure between two images in the probability space. Thus, using pairwise distances, we can take a core-set \cite{sener2018active} style approach for sample selection. Additionally, we can readily aggregate $ d_{\tiny[\bI_{m},\bI_{n}]} $ over the selected batch of images, $ \calI_b \subseteq \calI$ to compute the aggregate contextual diversity
\begin{equation}
\small
\label{eqn:CD}
d_{\tiny \calI_b} = \sum_{\tiny \bI_m, \bI_n \in \calI_b} d_{\tiny [\bI_{m},\bI_{n}]}.
\end{equation}
We use this term as the primary reward component in our RL framework. In addition to the intuitive motivation of using the contextual diversity, we show extensive comparisons in Sec. \ref{sec:expts} and ablative analysis in Sec. \ref{sec:analysis}. 

\subsection{Frame Selection Strategy}\label{sec:frame_sel}

\myfirstpara{CDAL-CS} 
Our first frame selection strategy is contextual diversity based active learning using core-set (CDAL-CS), which is inspired by the theoretically grounded core-set approach \cite{sener2018active}. To use core-set with Contextual Diversity, we simply replace the Euclidean distance with the pairwise contextual diversity (Eq. \ref{eqn:KL_div}) and use it in the K-Center-Greedy algorithm \cite[Algo. 1]{sener2018active}, which is reproduced in the supplementary material for completeness. 

\mypara{CDAL-RL} 
Reinforcement Learning has been used for frame selection \cite{Woodward_NIPSwDRL2016,Zhou_AAAI2018} for tasks like active one-shot learning and video summarization. We use contextual diversity as part of the reward function to learn a Bi-LSTM-based policy for frame selection. Our reward function comprises of the following three components. 

\mypara{Contextual Diversity ($ R_{cd} $)}
This is simply the aggregated contextual diversity, as given in Eq. (\ref{eqn:CD}), over the selected subset of images $ \calI_b $. 

\mypara{Visual Representation ($ R_{vr} $)} 
We use this reward to incorporate the visual representativeness over the whole unlabeled set using the image's feature representation. Let $ \bx_i $ and $ \bx_j $ be the feature representations of an image $ \bI_i\in\calI $ and of $ \bI_j\in\calI_b $ respectively, then
\begin{equation}
R_{vr} = \exp \left( -\frac{1}{\card{\calI}} \sum_{i=1}^{\card{\calI}} \min_{j \in {\footnotesize \calI_b}} \left( \norm{\bx_i - \bx_j}_2 \right) \right)
\end{equation}
This reward prefers to pick images that are spread out across the feature space, akin to $ k $-medoid approaches.

\mypara{Semantic Representation ($ R_{sr} $)}
We introduce this component to ensure that the selected subset of images are reasonably balanced across all the classes and define it as
\begin{equation}
    R_{sr} = \sum_{c\in C} \log \left( \card{ \calR_{\tiny\calI_b}^c} / \lambda \right) 
\end{equation} 
Here, $ \lambda $ is a hyper-parameter that is set to a value such that a selection that has substantially small representation of a class ($ \card{\calR^c_{\tiny\calI_b}} \ll \lambda $) gets penalized. We use this reward component only for the semantic segmentation application where certain classes (e.g., `pole') may occupy a relatively small number of regions (pixels). 

We define the total reward as $ R = \alpha R_{cd} + (1-\alpha)(R_{vr}+R_{sr}) $ and use it to train our LSTM based policy network. To emphasize the CD component in the reward function we set $ \alpha $ to $0.75$ across all tasks and experiments. The precise value of $ \alpha $ does not influence results significantly as shown by the ablation experiments reported in the supplementary. % {\color{red} Due to space limitations, we present the network architecture and training details, along with supporting ablation experiments in the supplementary material\footnote{The source code and additional results and analysis can be found on  \url{https://github.com/sharat29ag/CDAL}}.}% For completeness we include an ablation experiment in the supplementary material.  

%\begin{comment}
\subsection{Network Architecture and Training}
The contextual diversity measure is agnostic to the underlying \emph{task network} and is computed using the predicted softmax probability. Therefore in Sec. \ref{sec:expts}, our task network choice is driven by reporting a fair comparison with the state-of-the-art approaches on the respective applications. In the core-set approach \cite{sener2018active}, images are represented using the feature embeddings and pairwise distances are Euclidean. Contrarily, our representation is the mixture distribution computed in Eqn. (\ref{eqn:class_cont_conf}) over a single image and the corresponding distances are computed using pairwise contextual diversity in Eqn. (\ref{eqn:KL_div}).

For CDAL-RL, we follow a policy gradient based approach using the REINFORCE algorithm \cite{Williams_REINFORCE_1992} and learn a Bi-LSTM \emph{policy network}, where the reward used is as described in the previous section. The input to the policy network at a given time step is a representation of each image extracted using the task network. This representation is the vectorized form of an $ n_C\times n_C $ matrix, where the columns of the matrix are set to $ \bP^c_{\tiny\bI} $ for all $ c\in C $ such that $ \bI\in\calI^c $, and zero vectors otherwise. The binary action (select or not) for each frame is modeled as a Bernoulli random variable using a sigmoid activation at the output of the Bi-LSTM policy network. The LSTM cell size is fixed to 256 across all experiments with the exception of image classification, where we also show results with a cell size of 1024 to accommodate for a larger set of 100 classes. For REINFORCE, we use learning rate $ = 10^{-5} $, weight decay $ = 10^{-5} $, max epoch $ = 60 $ and \#episodes $ = 5 $. We achieve the best performance when we train the policy network from scratch in each iteration of AL, however, in Sec. \ref{sec:analysis} we also analyze and compare other alternatives. It is worth noting that in the AL setting, the redundancy within a large unlabeled pool may lead to multiple subsets that are equally good selections. CDAL-RL is no exception and multiple subsets may achieve the same reward, thus rendering the specific input image sequence to our Bi-LSTM policy network, irrelevant. % In addition to these details, we will publicly release the code with all training details upon acceptance. 
%\end{comment}

\section{Results and Comparison}\label{sec:expts}

We now present empirical evaluation of our approach on three visual recognition tasks of semantic segmentation, object detection and image classification\footnote{Additional results and ablative analysis is presented in the supplementary.}. % We also include analysis and ablation experiments for the semantic segmentation problem, and present additional ablations and qualitative results in the supplementary material.

\mypara{Datasets}
For semantic segmentation, we use Cityscapes \cite{cordts2016cityscapes} and BDD100K \cite{BDD100K_57_L}. Both these datasets have 19 object categories, with pixel-level annotation for 10K and 3475 frames for BDD100K and Cityscapes respectively. We report our comparisons using the mIoU metric. For direct comparisons with \cite{Yoo_2019_CVPR} over the object detection task, we combine the training and validation sets from PASCAL VOC 2007 and 2012 \cite{PASCAL_VOC_10_L} to obtain 16,551 unlabeled pool and evaluate the model performance on the test set of PASCAL VOC 2007 using the mAP metric. We evaluate the image classification task using classification accuracy as the metric over the CIFAR-10 and CIFAR-100 \cite{CIFAR10-100-29_V} datasets, each of which have 60K images evenly divided over 10 and 100 categories respectively.

\mypara{Compared Approaches}
The two recent works \cite{Yoo_2019_CVPR,VAAL_ICCV2019} showed state of the art AL performance on various visual recognition tasks and presented a comprehensive empirical comparison with prior work. We follow their experimental protocol for a fair comparison and present our results over all the three tasks. For the semantic segmentation task, we use the reported results for VAAL and its other competitors from \cite{VAAL_ICCV2019}, which are core-set \cite{sener2018active}, Query-by-Committee (QBC) \cite{Kuo_MICCAI2018}, MC-Dropout \cite{MCDropout_21_V} and Suggestive Annotation (SA) \cite{SA_55_V}. We refer to our contextual diversity based approaches as CDAL-CS for its core-set variant and CDAL-RL for the RL variant, which uses the combined reward $ R $ as defined in Sec.\ref{sec:frame_sel}. The object detection experiments are compared with learn loss \cite{Yoo_2019_CVPR} and its competitors -- core-set, entropy based and random sampling -- using results reported in \cite{Yoo_2019_CVPR}. For the image classification task, we again compare with VAAL, core-set, DBAL \cite{DBAL_16_V} and MC-Dropout. All the CDAL-RL results are reported after averaging over three independent runs. In Sec. \ref{sec:analysis} we demonstrate the strengths of $ CD $ through various ablative analysis on the Cityscapes dataset. Finally, in the supplementary material, we show further comparisons with region based approaches \cite{iiith,Mackowiak_CEREALS_BMVC2018}, following their experimental protocol on the Cityscapes dataset.   

\begin{figure}[h]
	\centering
	\begin{tabular}{cc}
		\includegraphics[width=0.35\linewidth]{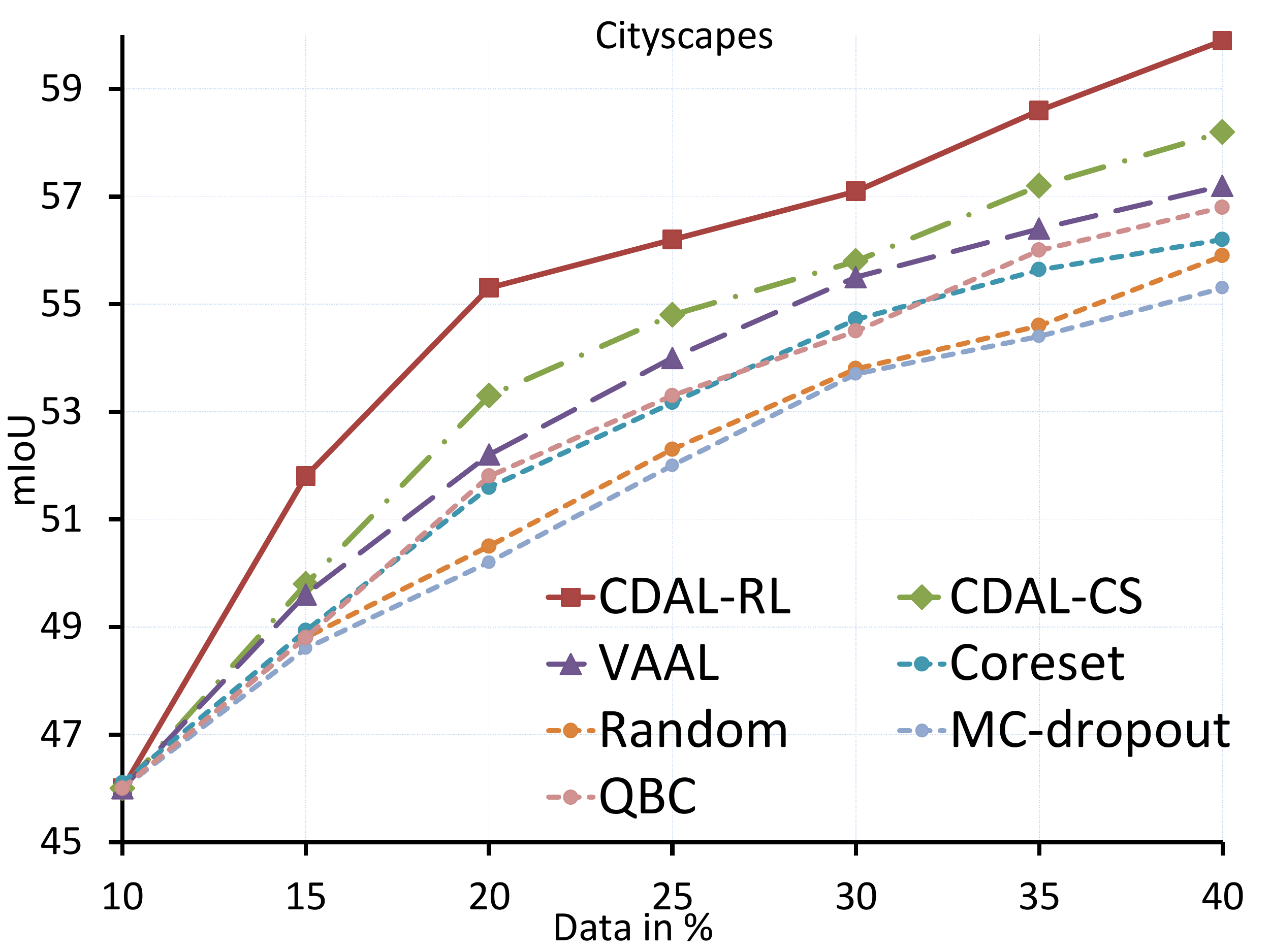} &
		\includegraphics[width=0.35\linewidth]{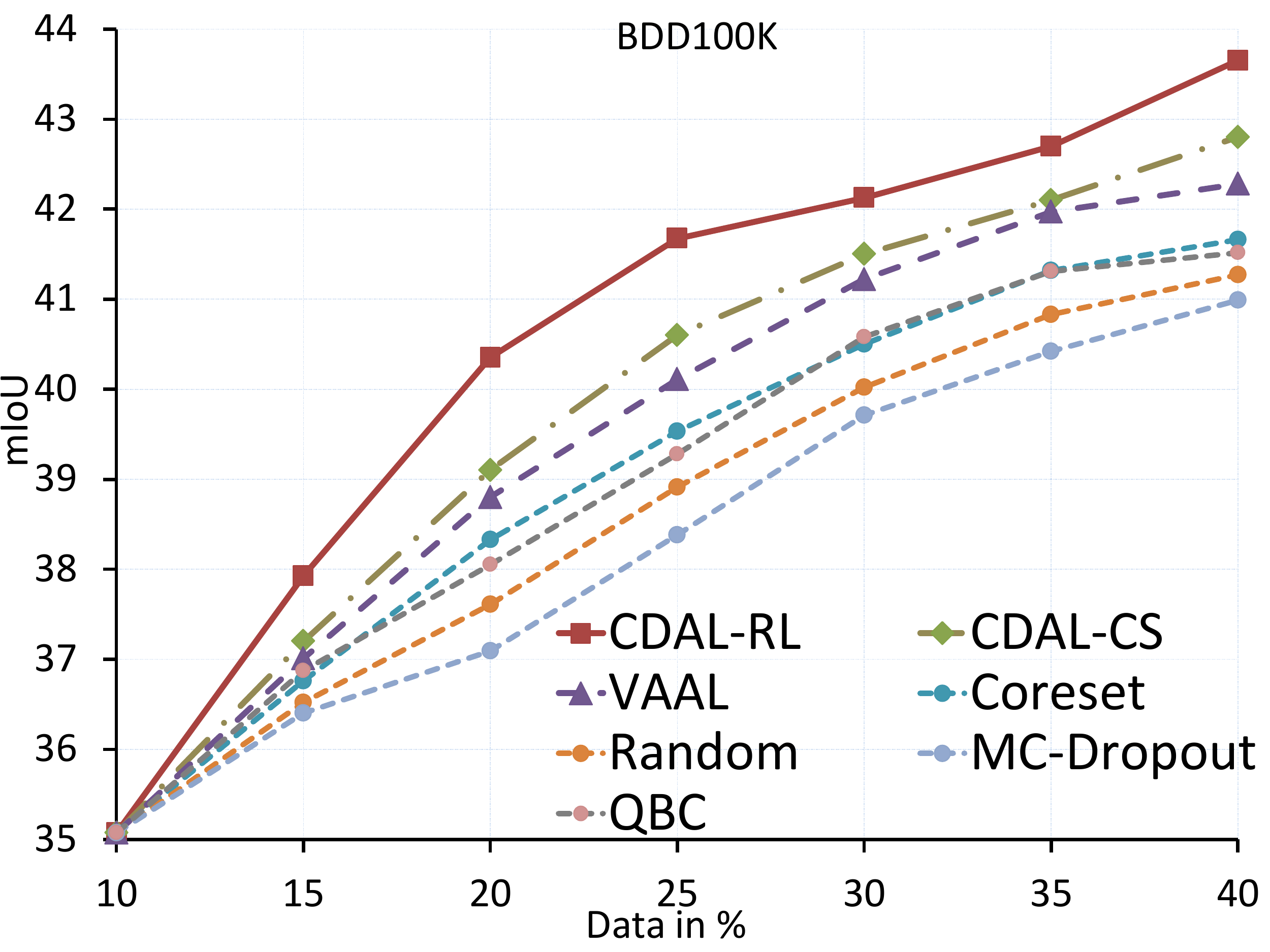}
		%(a) & (b)
	\end{tabular}
	\caption{Quantitative comparison for the Semantic Segmentation problem over Cityscapes \textbf{(left)} and BDD100K \textbf{(right)}. Note: DRN results 62.95\% and 44.95\% mIoU on 100\% data for Cityscapes and BDD respectively (best viewed in color).}
	\label{fig:Quant_seg}
\end{figure}

\subsection{Semantic Segmentation}
\label{sec:expts_seg}
Despite the tediousness associated with semantic segmentation, there are limited number of works for frame-level selection using AL. 
% The time taken to annotate an image for the semantic segmentation is arguably the highest among the visual recognition tasks. Yet we find very few competing frame-level selection AL techniques in the literature, possibly due to the complexity of the task. 
A recent approach applied to this task is VAAL \cite{VAAL_ICCV2019}, which achieves state-of-the-art performance while presenting a comprehensive comparison with previously proposed approaches. We follow the experimental protocol of VAAL \cite{VAAL_ICCV2019}, and also use the same backbone model of dilated residual network (DRN) \cite{DRN_56_V} for a fair comparison. As in their case, the annotation budget is set to 150 and 400 for Cityscapes and BDD100K respectively. The evaluation metric is mIoU. For each dataset, we evaluate the performance of an AL technique at each step, as the number of samples to be selected are increased from 10\% to 40\% of the unlabeled pool size, in steps of 5\%. Fig. \ref{fig:Quant_seg} shows the comparison over the two datasets. 

We observe that for both challenging benchmarks, the two variants of CDAL comprehensively outperform all other approaches by a significant margin. Our CDAL-RL approach can acheive current SOTA 57.2 and 42.8 mIoU by reducing the labeling effort by 300 and 800(10\%) frames on cityscapes and BDD100k respectively. A network's performance on this task is the most affected by the spatial context, due to the fine-grained spatial labeling necessary for improving the mIoU metric. We conclude that the CD measure effectively captures the spatial and semantic context and simultaneously helps select the most informative samples. 
There exist region-level AL approaches to semantic segmentation, where only certain regions are annotated in each frame \cite{iiith,Mackowiak_CEREALS_BMVC2018}. 
% An alternate approach to AL for semantic segmentation is to make region-level selections, choosing only specific `most informative' regions from each frame in the dataset \cite{iiith,Mackowiak_CEREALS_BMVC2018}. % While a direct comparison with frame-level selection based methods may not be fair, we follow the experimental settings provided in the respective papers and report a comparative analysis in the supplementary material. 
Our empirical analysis in the supplementary material shows that our CDAL based frame selection strategy is complementary to the region-based approaches.

\subsection{Object Detection} 

% \begin{comment}
\begin{wrapfigure}[17]{r}{0.45\textwidth}
%\begin{figure}
	\centering
	\includegraphics[width=0.85\linewidth]{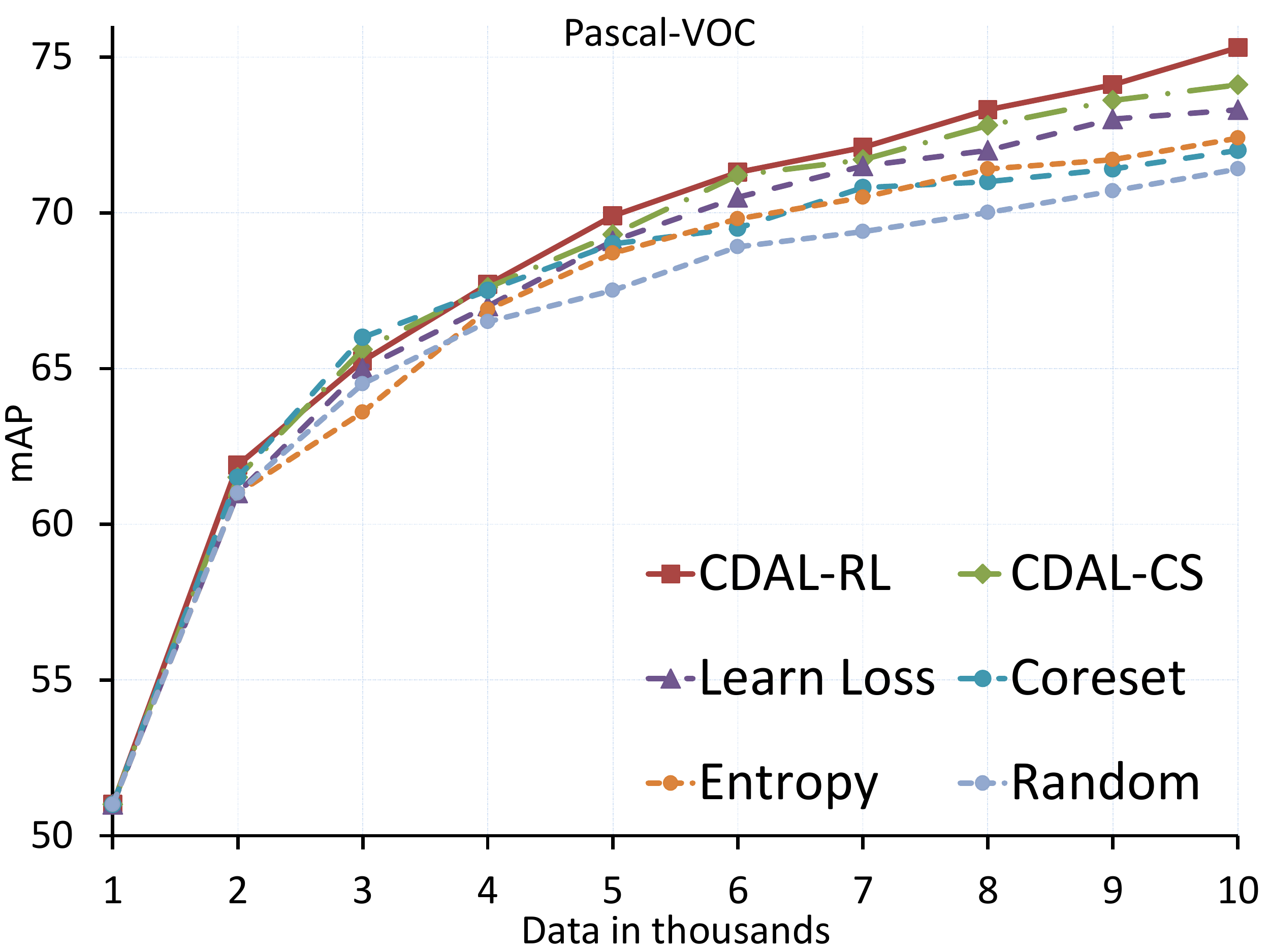}
	\caption{Quantitative comparison for Object Detection over PASCAL-VOC dataset. We follow the experimental protocol of the learning loss method \cite{Yoo_2019_CVPR}.  %The numbers reported for the compared methods are also as given in \cite{Yoo_2019_CVPR}.
	Note: SSD results 77.43\% mAP on 100\% data of PASCAL-VOC(07+12).}
	\label{fig:Quant_det}
%\end{figure}
\end{wrapfigure}
%\end{comment}	
For the object detection task, we compare with the learning loss approach \cite{Yoo_2019_CVPR} and the competing methods therein. % by following the experimental protocol and mAP values reported in their paper. The other competing methods presented for comparisons here are baselines like random sampling and entropy based uncertainty sampling, and core-set \cite{sener2018active}.
For a fair comparison, we use the same base detector network as SSD \cite{SSD_ECCV2016} with a VGG-16 \cite{simonyan2014very} backbone and use the same hyperparameter settings as described in \cite{Yoo_2019_CVPR}. % We use the mAP metric for evaluation and the dataset used is the combined Pascal-VOC as described earlier. 

Fig. \ref{fig:Quant_det} shows the comparisons, where we see in most cases, both variants of CDAL perform better than the other approaches. During the first few cycles of active learning, i.e., until about 5K training samples are selected for annotation, CDAL performs nearly as well as core-set, which outperforms all the other approaches. In the later half of the active learning cycles with 5K to 10K selected samples, CDAL variants outperform all the other approaches including core-set \cite{sener2018active}. CDAL-RL achieved 73.3 mAP using 8k data where learning loss \cite{Yoo_2019_CVPR} achieved it by 10k hence reducing 2k labeled samples. 

\begin{figure*}[t!]
	\centering
	\setlength{\tabcolsep}{-1pt}
	\begin{tabular}{cc}
		\includegraphics[width=0.35\linewidth]{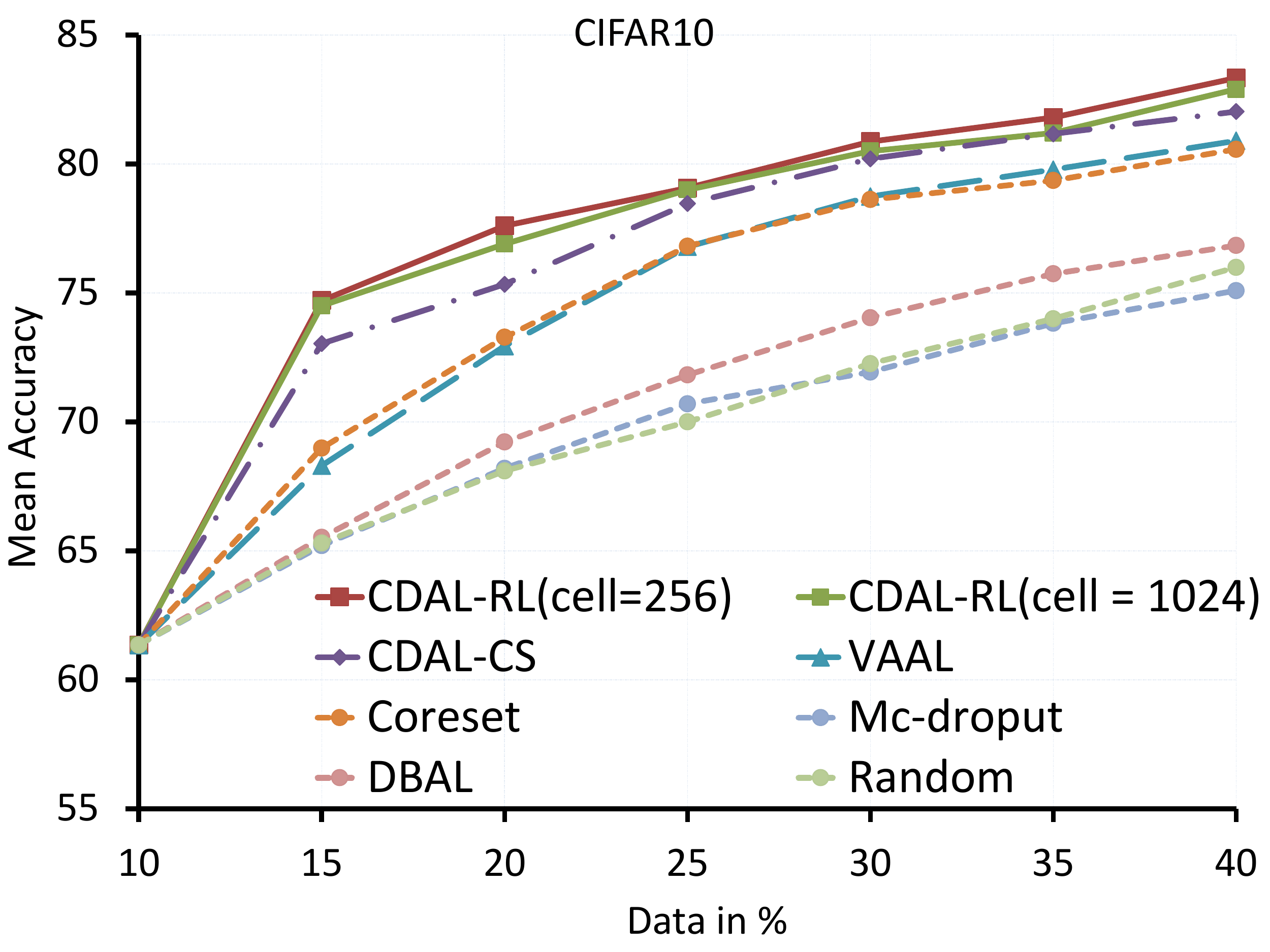} &
		\includegraphics[width=0.35\linewidth]{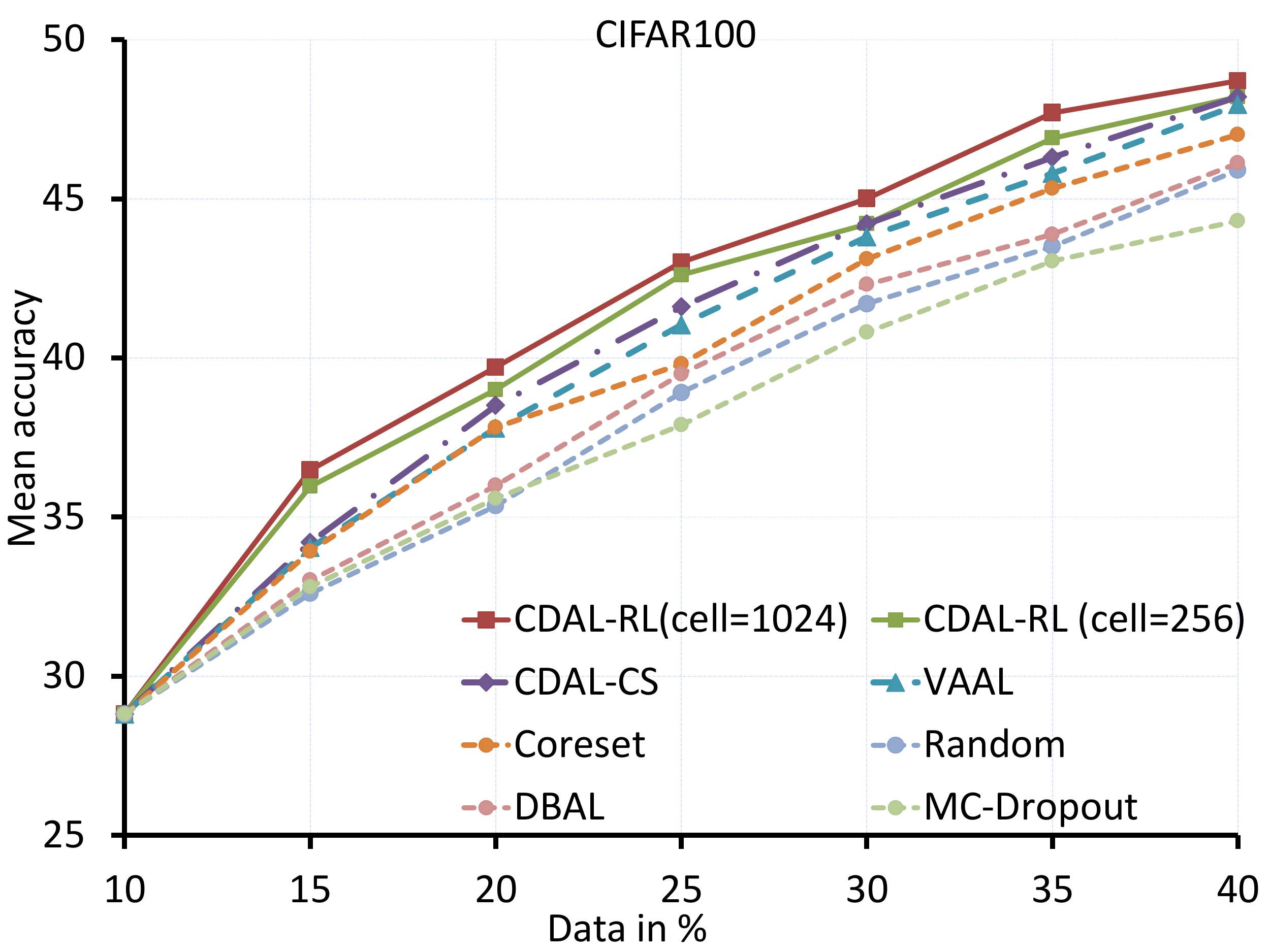}
		%(a) & (b)
	\end{tabular}
	\caption{Quantitative comparison for Image Classification over CIFAR-10 \textbf{(left)} and CIFAR-100 \textbf{(right)}. CDAL-RL(cell=$n$) indicates that the LSTM policy network has a cell size $ n $. Note: VGG16 results 90.2\% and 63.14\% accuracy on 100\% data for CIFAR10 and CIFAR100 respectively (best viewed in color).}
	\label{fig:Quant_class_det}
\end{figure*}

\subsection{Image Classification}
One of the criticisms often made about the active learning techniques is their relative difficulty in scaling with the number of classes in a given task. For example it has been reported in \cite{VAAL_ICCV2019}, that core-set \cite{sener2018active} does not scale well for large number of classes. To demonstrate the strength of contextual diversity cues when the number of classes is large, we present the evaluations on the image classification task using CIFAR-10 and CIFAR-100. Fig. \ref{fig:Quant_class_det} shows the comparison. It is clear that CDAL convincingly outperforms the state of the art technique, VAAL \cite{VAAL_ICCV2019} on both the datasets. We can see that CDAL-RL can achieve $\sim$81\% accuracy on CIFAR10 by using 5000 (10\%) less samples than VAAL \cite{VAAL_ICCV2019} and similarly 2500 less samples are required in CIFAR100 to beat SOTA of 47.95\% accuracy. These results indicate that CDAL can gracefully scale with the number of classes, which is not surprising as CD is a measure computed by accumulating KL-divergence, which scales well with high-dimensions unlike the Euclidean distance. It is worth noting that an increase in the LSTM cell size to 1024, helps improve the performance on CIFAR-100, without any significant effect on the CIFAR-10 performance. A higher dimension of the LSTM cell has higher capacity which better accommodates a larger number of classes. For completeness, we include more ablations of CDAL for image classification in the supplementary.

We also point out that in image classification, the entire image qualifies as a \emph{region} (as defined in Sec. \ref{sec:CDAFSN}), and the resulting mixture $ \bP^c_{\tiny \bI} $ comprising of a single component still captures confusion arising from the spatial context. Therefore, when a batch $ \calI_b $ is selected using the contextual diversity measure, the selection is diverse in terms of classes and their confusion. 

\section{Analysis and Ablation Experiments}
\label{sec:analysis}
In the previous section, we showed that contextual diversity consistently outperforms state of the art active learning approaches over three different visual recognition tasks. We will now show a series of ablation experiments to demonstrate the value of contextual diversity in an active learning problem. Since active learning is expected to be the most useful for the semantic segmentation task with highest amount of annotation time per image, we have chosen the task for our ablation experiments. We have designed all our ablation experiments on the Cityscapes dataset using the DRN model in the same settings as in Sec. \ref{sec:expts_seg}. 
%\subsection{CDAL-RL Analysis}
\begin{figure}[t!]
	\centering
	\begin{tabular}{cc}
		\includegraphics[width=0.35\linewidth]{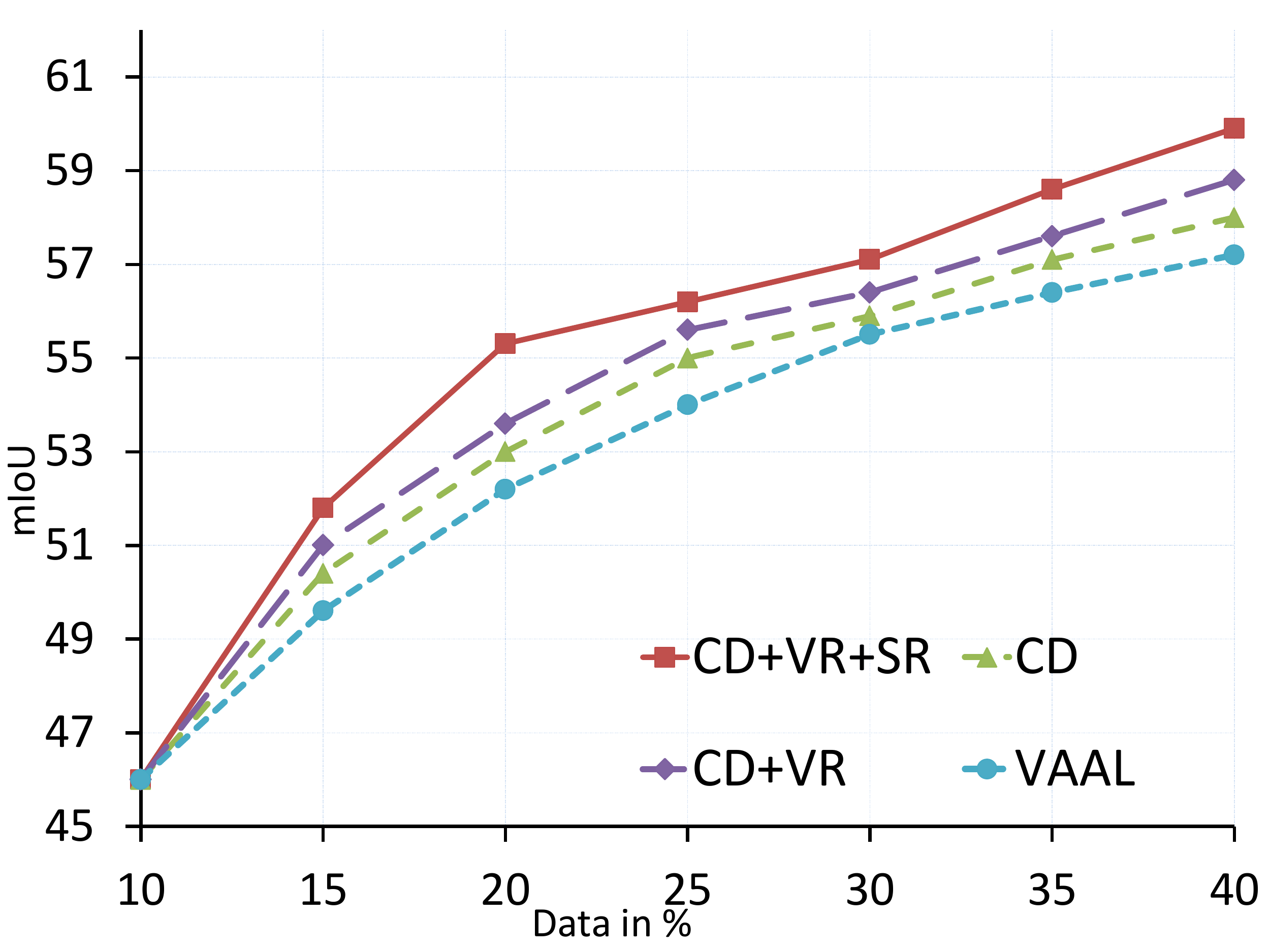}&
		\includegraphics[width=0.35\linewidth]{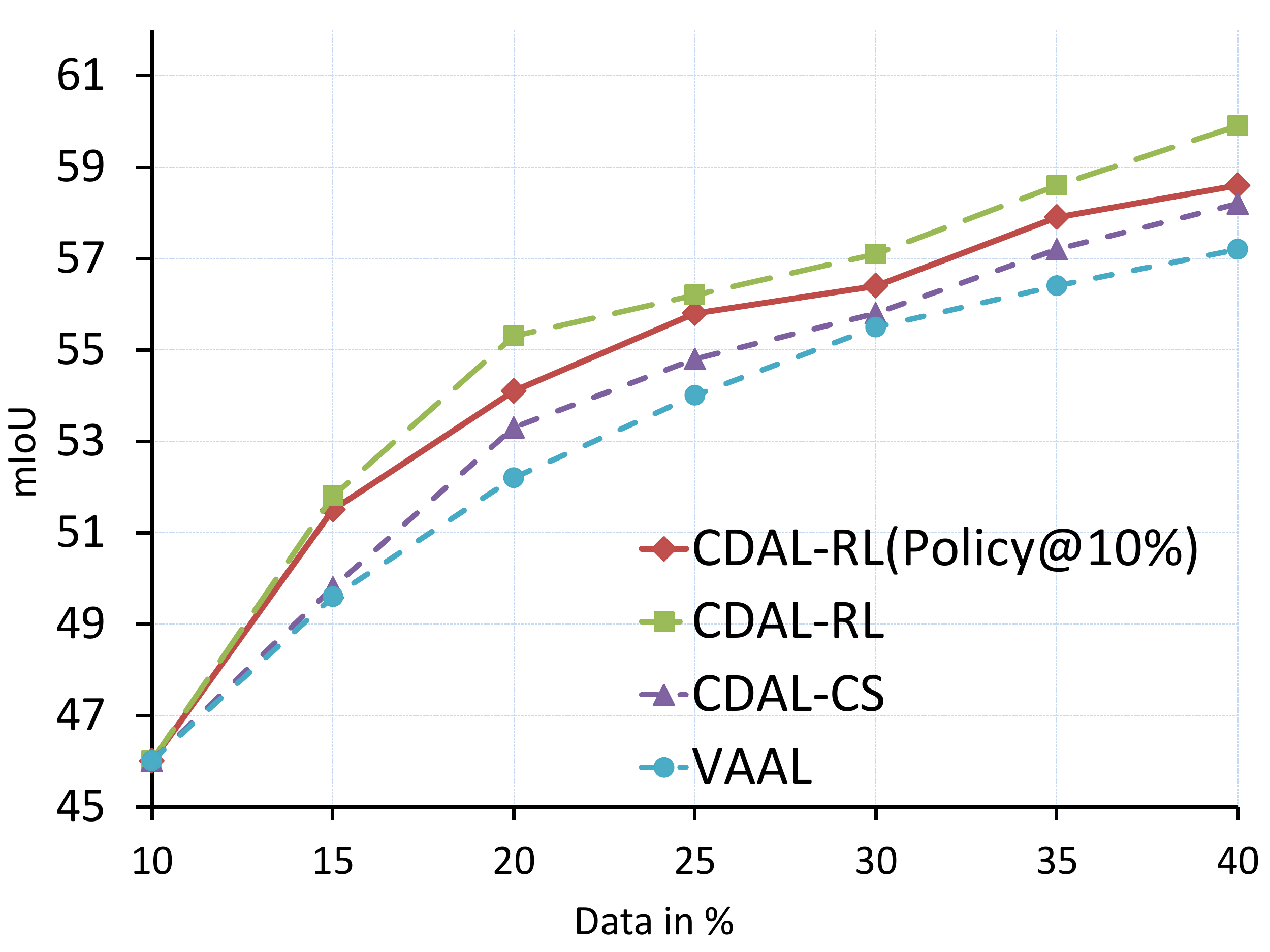}% \\
%		(a) & (b)
	\end{tabular}
	\caption{\textbf{(left)} Ablation with individual reward components on Cityscapes.  \textbf{(right)} Cityscapes results when the CDAL-RL policy was learned only once in the first iteration with 10\% randomly selected frames (best viewed in color).}
	\label{fig:all_reward_ablation}
\end{figure}

\noindent\textbf{Reward Component Ablation}. We first investigate the performance of various components of the reward used in our approach. Fig. \ref{fig:all_reward_ablation}(left) shows the performance of CDAL in three different reward settings: only contextual diversity ($ R=R_{cd} $), contextual diversity and visual representation (CD+VR, i.e., $R= \alpha R_{cd}+(1-\alpha)R_{vr} $) and all the three components including semantic representations (CD+VR+SR, i.e., $ R=\alpha R_{cd}+(1-\alpha)(R_{vr}+R_{sr}) $). It is clear that contextual diversity alone outperforms the state of the art, VAAL \cite{VAAL_ICCV2019}, and improves further when the other two components are added to the reward function. As mentioned in Sec. \ref{sec:frame_sel}, the value of $ \alpha =0.75$ was not picked carefully, but only to emphasize the CD component, and remains fixed in all experiments. 

\noindent\textbf{Policy Training Analysis.} Our next experiment analyzes the effect of learning the Bi-LSTM-based policy only once, in the first AL iteration. We train the policy network using the \emph{randomly selected} 10\% and use it in each of the AL iterations for frame selection without further fine-tuning. The results are shown in Fig. \ref{fig:all_reward_ablation}(right), where we can see that this policy denoted by CDAL-RL(policy@10\%), still outperforms VAAL and CDAL-CS in \emph{all} iterations of AL. Here CDAL-RL is the policy learned under the setting in Sec. \ref{sec:expts_seg}, where the policy network is trained from scratch in each AL iteration. An interesting observation is the suitability of the contextual diversity measure as a reward, and that it led to learning a meaningful policy even with randomly selected data. 

\begin{figure*}[t!]
	\centering
	\begin{tabular}{cc}
		\includegraphics[width=0.4\linewidth]{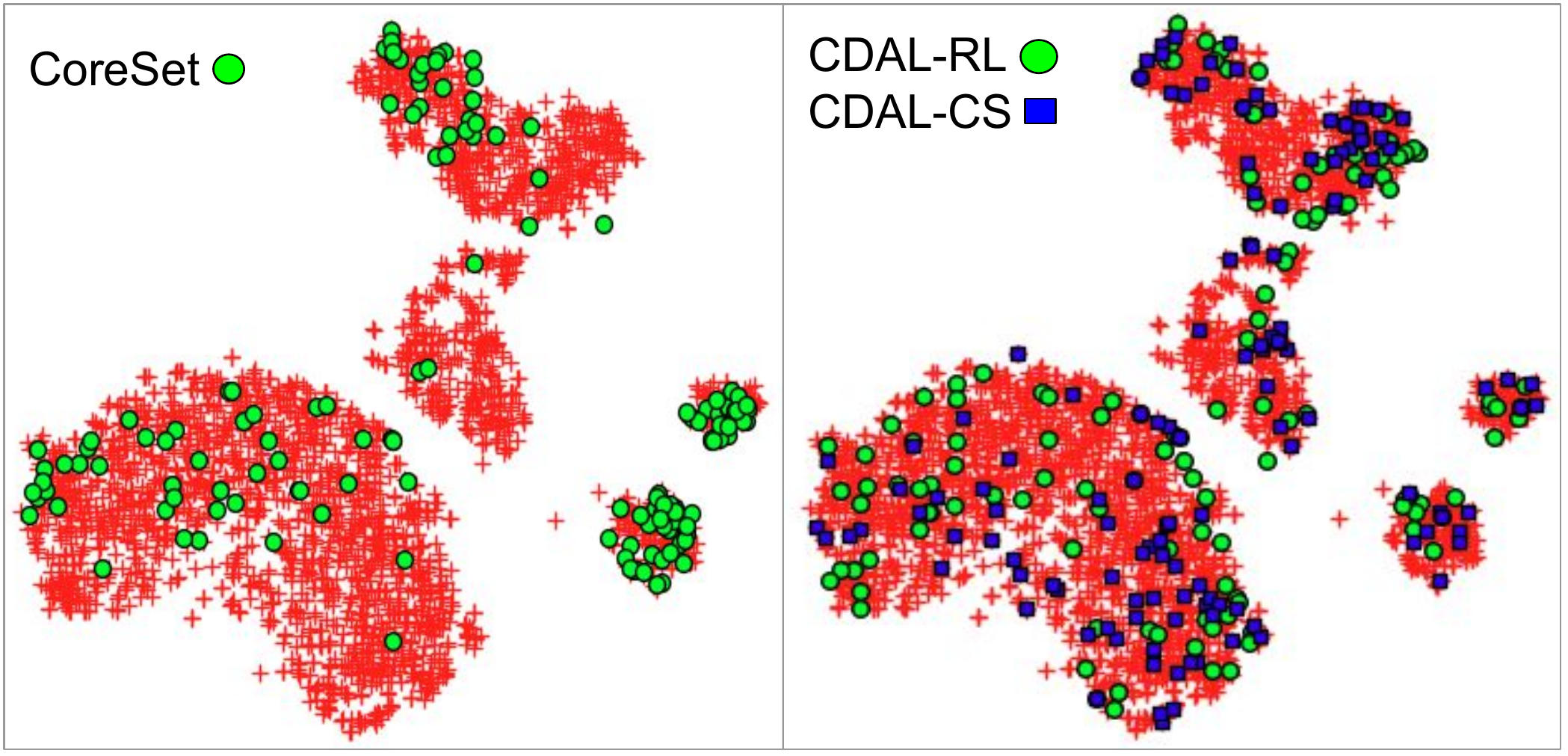} &
		\includegraphics[width=0.4\linewidth]{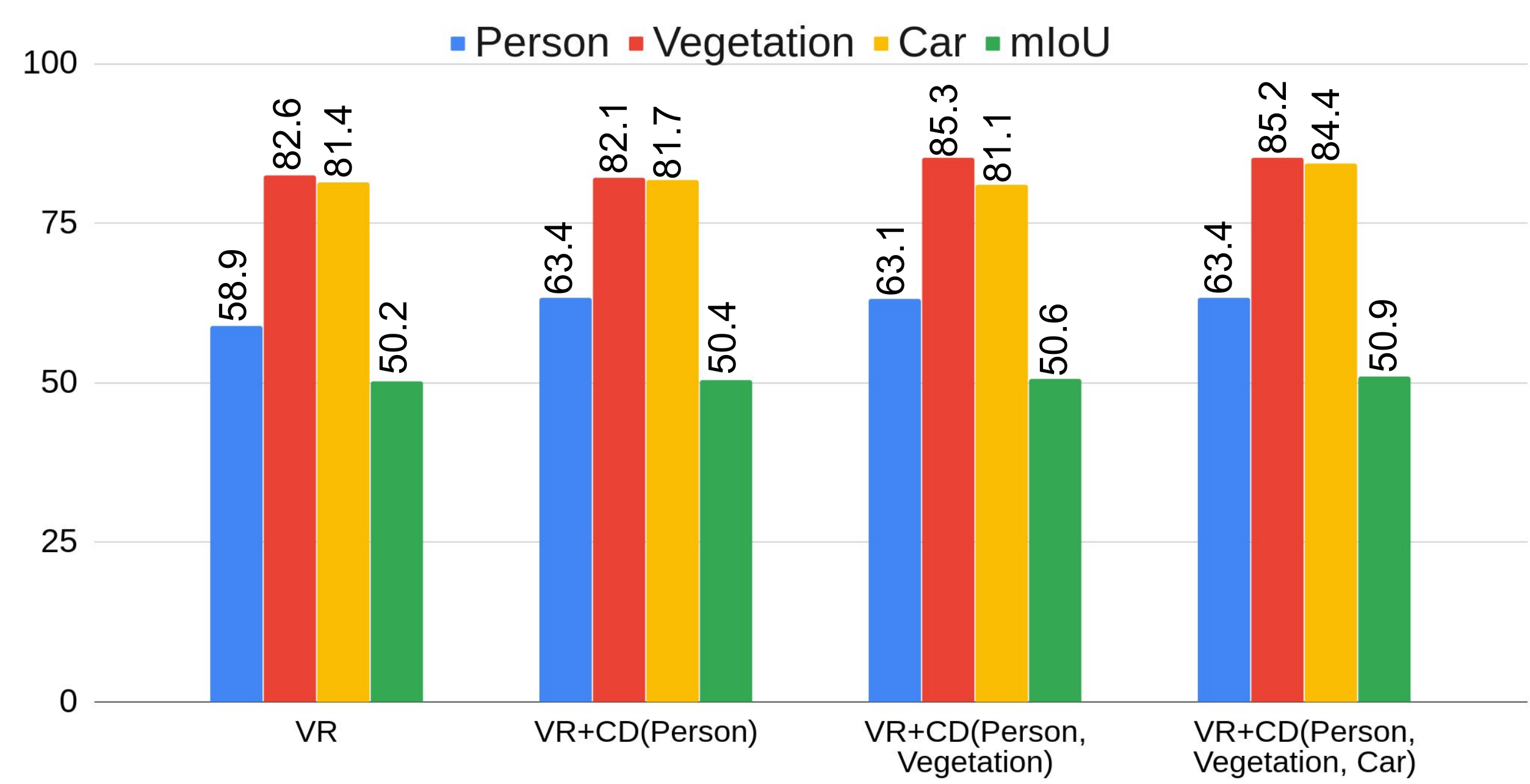}	\\
		(a) & (b)
	\end{tabular}
	\caption{(a) t-sne plots comparison with \cite{sener2018active} on Cityscapes: CoreSet (left), and CDAL (right) (b) Performance analysis when CD reward is computed for an increasing number of classes.(best viewed in color)}
	\label{fig:t_sne_cdananlysis_ablation}
\end{figure*}

\noindent\textbf{Visualization of CDAL-based Selection.} In Fig. \ref{fig:t_sne_cdananlysis_ablation}(a), we show t-SNE plots \cite{tsne_JMLR2008} to visually compare the distribution of the points selected by the CDAL variants and that of core-set. We use the Cityscapes training samples projected into the feature space of the DRN model. The red points in the plots show the unlabeled samples. The left plot shows green points as samples selected by core-set and the right plot shows green and blue points are selected by CDAL-RL and CDAL-CS respectively. It is clear that both variants of the contextual diversity based selection have better spread across the feature space when compared with the core-set approach, which is possibly due to the distance concentration effect as pointed by \cite{VAAL_ICCV2019}. This confirms the limitation of the Euclidean distance in a high-dimensional feature space corresponding to the DRN model. On the other hand, CDAL selects points that are more uniformly distributed across the entire feature space, reflecting better representativeness capability. 

\noindent\textbf{Class-wise Contextual Diversity Reward.} The CD is computed by accumulating the symmetric KL-divergence (cf. Eq. (\ref{eqn:KL_div})) over all classes. Therefore, it is possible to use the $ R_{cd} $ reward only for a few, and not all, the classes. Fig. \ref{fig:t_sne_cdananlysis_ablation}(b) shows the segmentation performance as we incorporate the contextual diversity (CD) from zero to three classes. The initial model is trained using only the visual representation reward ($ R_{vr} $) and is shown as the leftmost group of color-coded bars. As we include the $ R_{cd} $ term in the reward with the CD only being computed for the \emph{Person} class, we see a substantial rise in the IoU score corresponding to \emph{Person}, as well as a marginal overall improvement. As expected, when we include both, the \emph{Person} and \emph{Vegetation} classes in the CD reward, we see substantial improvements in both the classes. %However, using CD reward only for a few classes, may bias the frame selection in favor of the chosen classes, and negatively affect the performance of classes not included in the CD term. The effect may be especially pronounced for the less frequent and the classes uncorrelated with the chosen classes. For example, the IoU for the \emph{Vegetation} class reduces when only the \emph{Person} class was used, but the performance of the \emph{Car} class was not affected. 
The analysis indicates that the contextual diversity reward indeed helps mitigating class-specific confusion.

\noindent\textbf{Limitations of CDAL.} While we show competitive performance of the policy without retraining (Fig.\ref{fig:all_reward_ablation}(right)), for best performance retraining at each AL iteration is preferred. For large datasets, this requires larger unrolling of the LSTM network incurring more computational and memory costs. Another limitation of CDAL in general is in the case of image classification, where the entire image is treated as a single region and thus is unable to fully leverage spatial context. 

\section{Conclusion}\label{sec:conc}
We have introduced a novel contextual diversity based measure for active frame selection. Through experiments for three visual recognition tasks, we showed that contextual diversity when used as a distance measure with core-set, or as a reward function for RL, is a befitting choice. It is designed using an information-theoretic distance measure, computed over the mixture of softmax distributions of pseudo-labeled data points, which allows it to capture the model's predictive uncertainty as well as class-specific confusion. We have only elaborated the promising empirical results in this paper, and plan to investigate deeper theoretical interpretations of contextual diversity that may exist.

\subsection*{Acknowledgement}
The authors acknowledge the partial support received from the Infosys Center for Artificial Intelligence at IIIT-Delhi. This work has also been partly supported by the funding received from DST through the IMPRINT program (IMP/2019/000250).

\bibliographystyle{splncs04}
\bibliography{CDAL_main}

\newpage
{\Large{\centering \textbf{Supplementary Material}}}

\section{Region Level Selection} 

We also compare with two approaches that are primarily considered as a region level selection approach: CEREALS \cite{Mackowiak_CEREALS_BMVC2018} and RBAL \cite{iiith}. CEREALS uses an FCN-8s \cite{Long_2015_CVPR} (with a 0.25 width multiplier) architecture, while RBAL uses the ICNet \cite{zhao2018icnet} architecture. We use the respective architectures to compute the contextual diversity (CD) for frame selection in our experimental comparisons. CEREALS operates in different modes, one of which uses the entire frame as selected region. This setting, while not the best performing version of CEREALS, is directly comparable with CDAL. We report this comparison in the first row of Table \ref{tab:Quant_seg_region}. For this comparison, we follow the same protocol as presented in \cite{Mackowiak_CEREALS_BMVC2018} and take the initial seed set of 50 images and at each step another 50 samples are selected. We report the results when nearly 10\% of the data ise selected, which amounts to about 300 frames in Cityscapes. We can see that for frame selection CDAL-RL outperforms CEREALS by about 3.4\% mIoU. We also note that CDAL is complementary to the approach that CEREALS. The best configuration of CEREALS, which upon leveraging only small patches ($ 128\times 128 $) within an image achieves an mIoU of 57.5\% by annotating about 10\% of the data. 

In RBAL \cite{iiith}, the ICNet \cite{zhao2018icnet} model was pre-trained using 1175 frames, followed by selection of 10\% of pixels across the remaining 1800 frames. For a fair comparison, we maintain the same annotation budget (1175 + 0.1*1800) of 1355 frames. We pre-train the ICNet model using 586 frames and select the remaining 769 frames using CDAL. The results after fine-tuning the model with the selected frames is compared with the mIoU reported in \cite{iiith} in the second row of Table \ref{tab:Quant_seg_region}. We observe a reasonable improvement, even though CDAL only selects 1355 frames as opposed to RBAL accessing all 2975 frames. 

Both of these results indicate that CDAL based frame selection complements the region based selection. 

\setlength{\tabcolsep}{5pt}
\begin{table}[h!]
\centering	
{\small
\begin{tabular}{lcccc}
	\toprule[1.5pt]
	& Selected  & Selected & Base & \multirow{2}{*}{mIoU (\%)} \\ 
	& Data & Frames & Model &  \\ 
	\toprule[1pt]
	CEREALS \cite{Mackowiak_CEREALS_BMVC2018}  & \multirow{2}{*}{$ \sim10 $\%} & 300 & \multirow{2}{*}{FCN8s} & 48.2   \\
	\textbf{CDAL-RL} & &  300 &  & \textbf{51.6}\\
	\midrule[0.5pt]
	RBAL \cite{iiith}& \multirow{2}{*}{45\%} & 2975 & \multirow{2}{*}{ICNet} & 61.3  \\
	\textbf{CDAL-RL}& & 1355 & & \textbf{62.9}  \\
	%\midrule[0.5pt]
	%\textbf{CDAL-RL + RB} & 10.5\% & $ 313^\ast $ & FCN8s & \textbf{58.1}\\
	\bottomrule[1.5pt] 
\end{tabular}
}
	\caption{Comparisons with region-based active learning approaches on Cityscapes. CDAL-RL again outperforms both the methods with a significant margin.% $ ^\ast $ CDAL-RL when combined with uncertainty based region selection, effectively annotating only ~10.5\% of pixels ($ \sim  $313 frames), it still achieves an mIoU higher than CEREALS.
	}
	\label{tab:Quant_seg_region}
\end{table}

\section{Qualitative Results for CDAL}

\begin{figure}[h!]
	\centering
	\begin{tabular}{c}
		\includegraphics[width=0.85\linewidth]{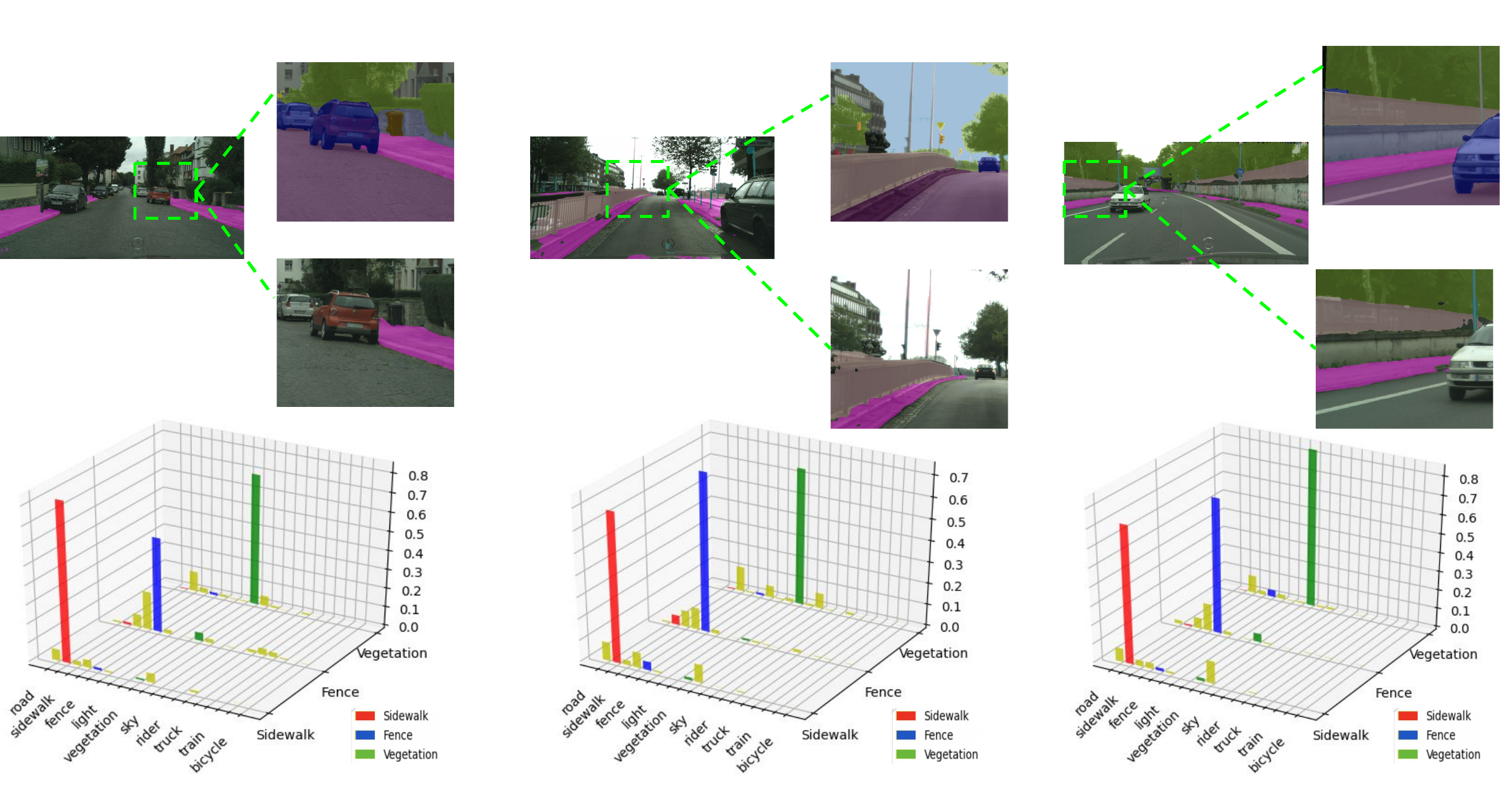}\\
		\includegraphics[width=0.85\linewidth]{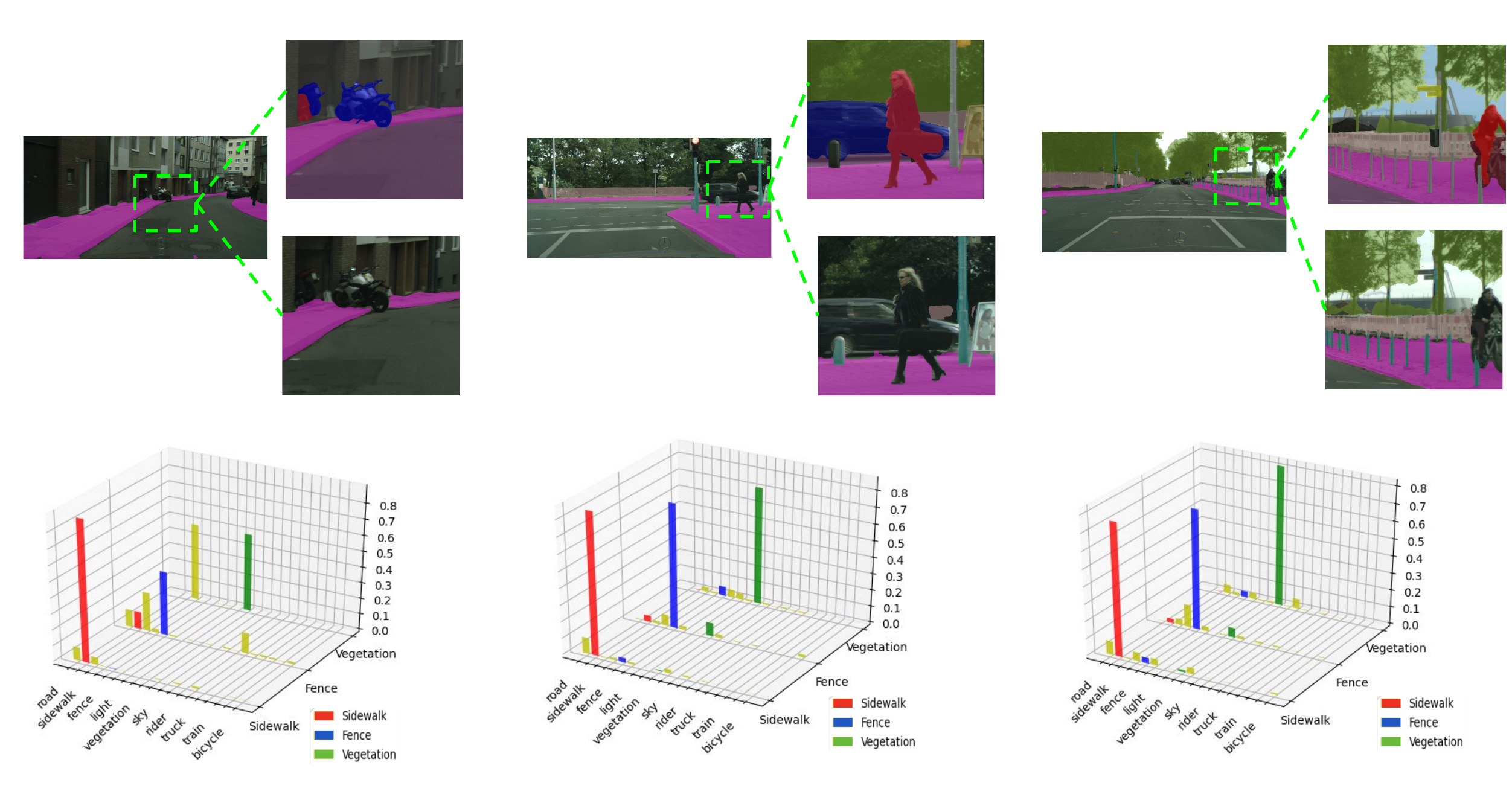}\\
		\includegraphics[width=0.85\linewidth]{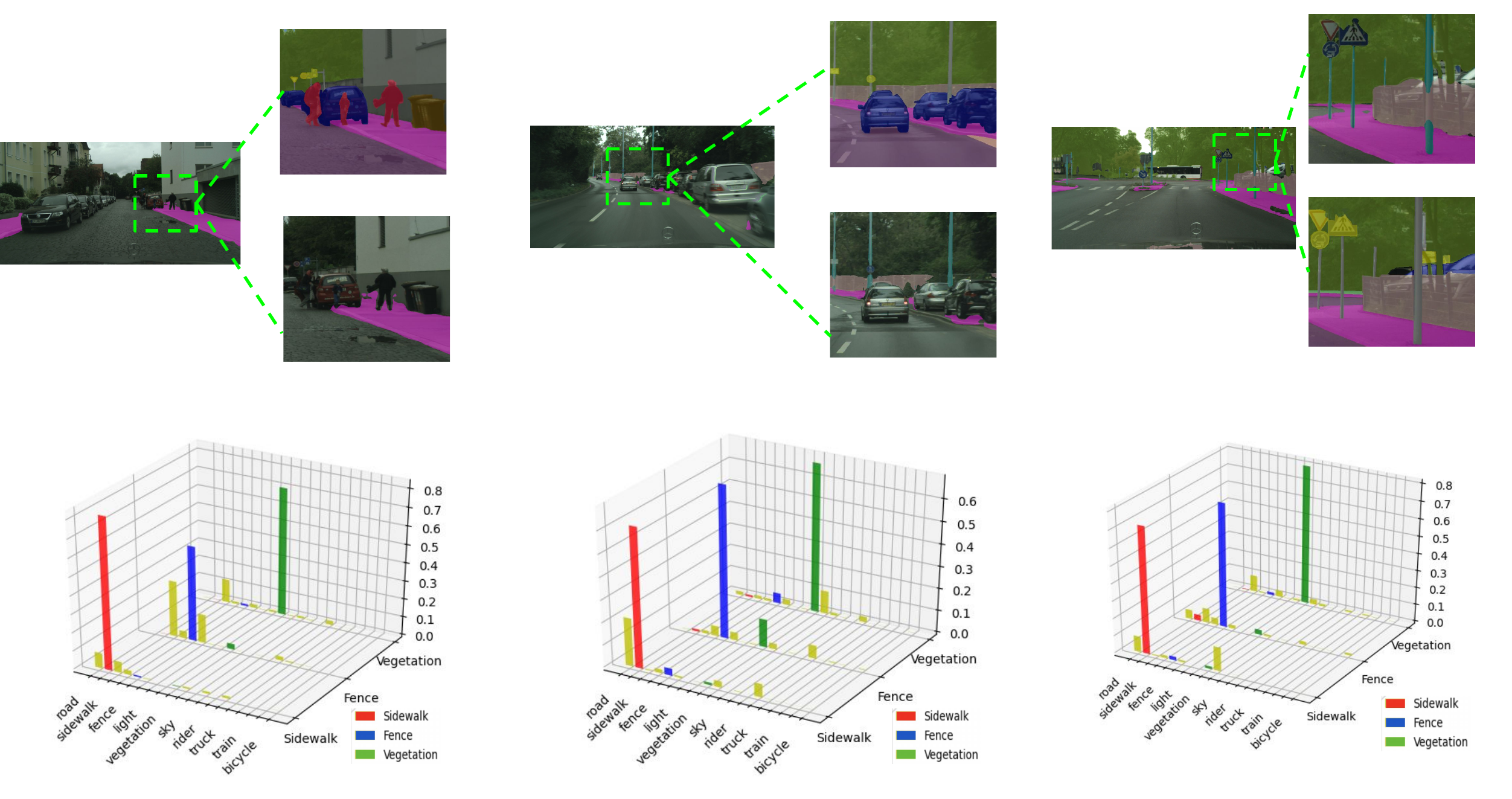}	
	\end{tabular}
	\caption{\textbf{Qualitative results for Class-wise Contextual Diversity Reward}. Columns (Left to Right) show outcomes of experiments run when the Contextual Diversity (CD) is computed using the following sets of classes: \{Sidewalk\}, \{Sidewalk and Fence\}, \{Sidewalk, Fence and Vegetation\}. Pairs of rows show frames selected by CDAL when CD was computed using the aforementioned classes in the top row. The top and bottom insets show the ground truth and the predictions overlaid over the regions respectively. The bottom row, shows the class-specific confusion corresponding to the three classes used for computing CD. The mixture distribution depicting the class-specific confusion is computed over the selected image. As more classes are included in the CD computation, we see the confusion reducing.}
	\label{fig:qual_cdal}
\end{figure}

In Fig. \ref{fig:qual_cdal}, we show the top 3 frames selected from three independent runs of CDAL-RL from the Class-wise Contextual Diversity Reward ablation described in Sec. 5 of the main paper. The left most column shows images selected when CDAL-RL is only trained with a CD reward using the set of classes \{Sidewalk\}, the middle column with CD computed using \{Sidewalk, Fence\} and the rightmost column with CD computed using \{Sidewalk, Fence, Vegetation\}. Two rows are shown for each selection. The first row shows the frame selected with the predictions for each class used for CD computation overlaid. The insets show the ground truth labels (top) and the predicted labels. The second row shows the class-specific confusion for the three classes considered: Sidewalk, Fence and Vegetation. 

As we scan through the rows, we see that as the classes are included in the CD computation, the corresponding class-confusion reduces, which is evident from the reduced entropy (and increased \emph{peakiness}) of each of the mixture distributions. As we see the selected images along the columns, we observe that the spatial neighborhood of the classes like Sidewalk contain different classes like Car, Motorcycle, and Person in the first column. Similarly, in the other two columns we see a different set of classes appearing in the selected frames in the spatial neighborhood of regions corresponding to the predictions of Sidewalk, Fence and Vegetation\footnote{More qualitative results are included in our project page at: \url{https://github.com/sharat29ag/CDAL}}. 

\section{Ablation on Image Classification: CIFAR100}
\begin{figure}
	\centering
	\begin{tabular}{cc}
		\includegraphics[width=0.5\linewidth]{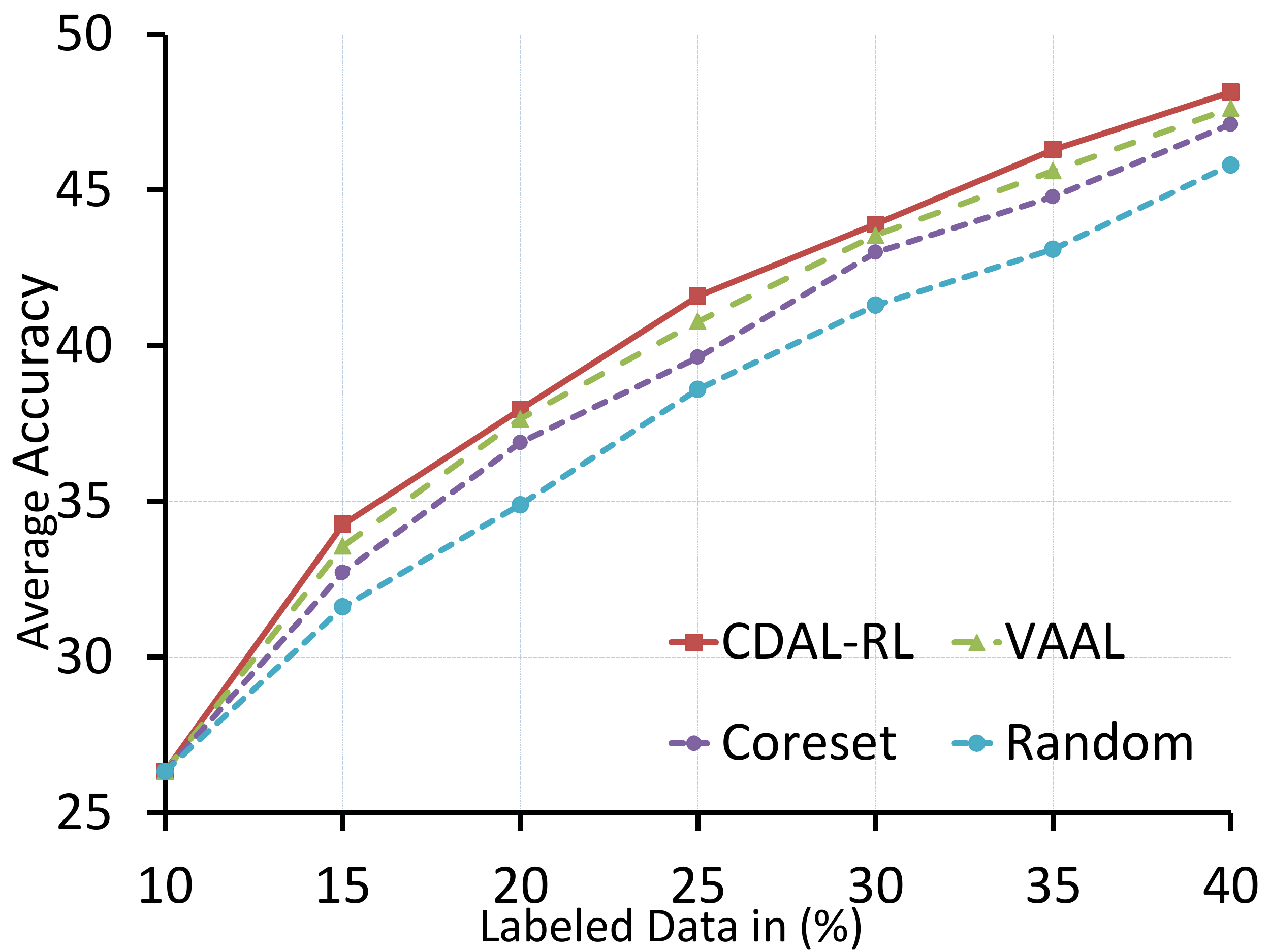} &
		\includegraphics[width=0.5\linewidth]{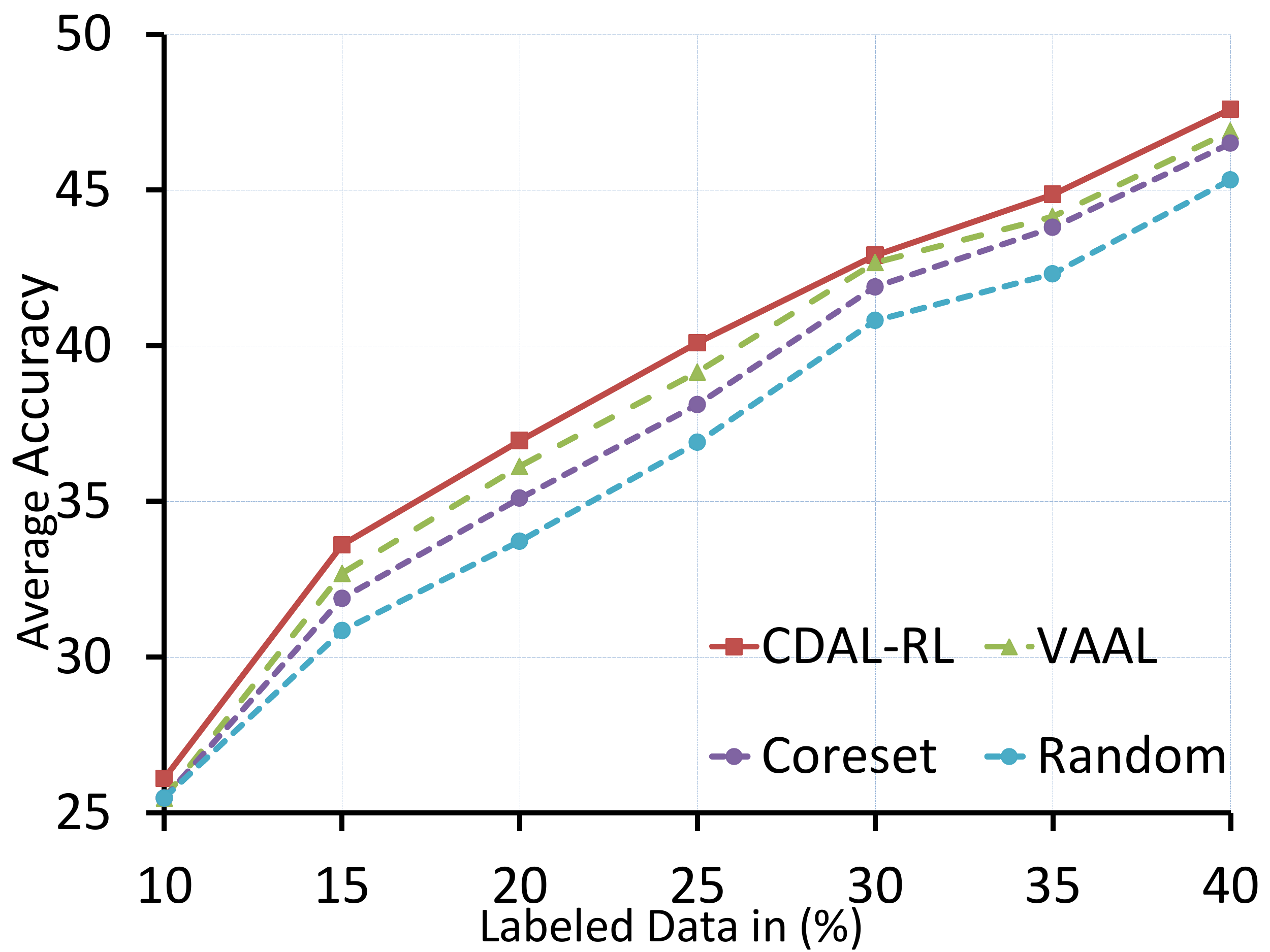}\\
		(a)&(b)\\
		\includegraphics[width=0.5\linewidth]{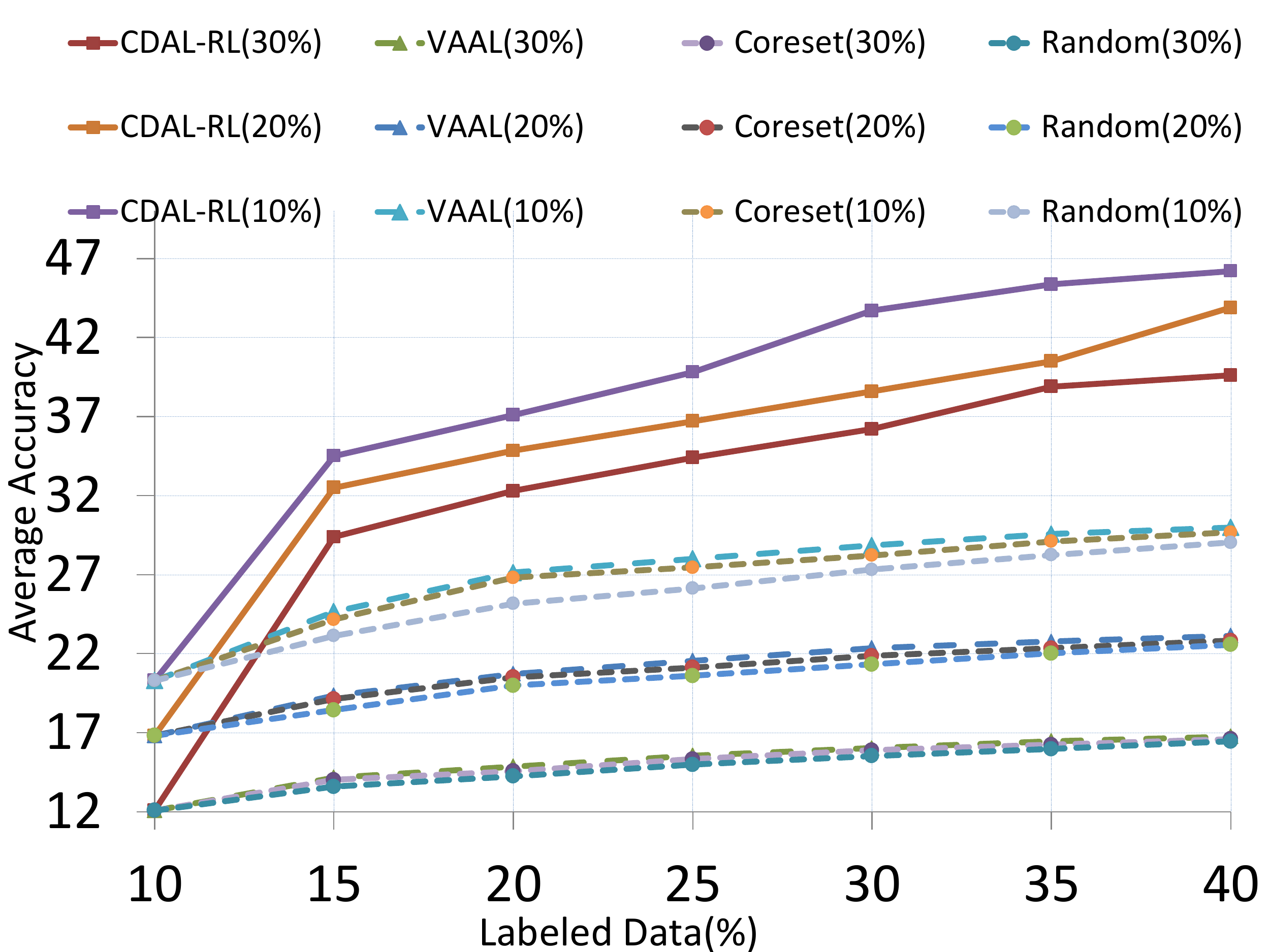} &
		\includegraphics[width=0.5\linewidth]{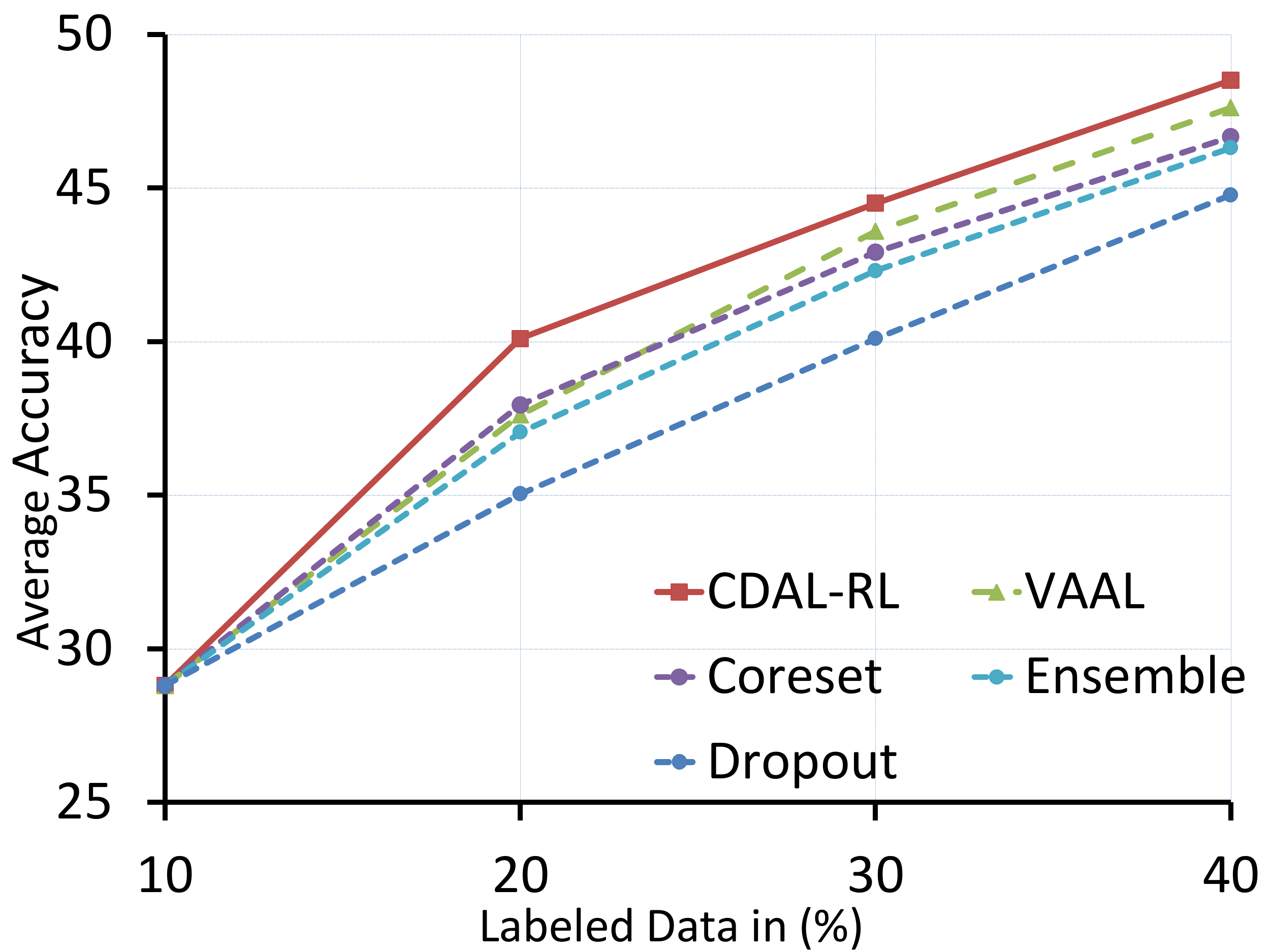}\\
		(c)&(d)\\
	\end{tabular}
	\caption{Ablation of CDAL-RL on CIFAR-100. Biased Initial pool with with (a)m=10 and (b)m=20, (c) Noisy Oracle, (d) Varying budget with a step size of 10\%}
	\label{fig:img_clsf_ablation}
\end{figure}

\mypara{Biased Initial Pool}
In the first case we check the robustness of CDAL in terms of biased initial labeled pool. We follow the exact experimental setup of VAAL \cite{VAAL_ICCV2019} and at random exclude data for m=10 and m=20 classes from the initial labeled pool. From Figures \ref{fig:img_clsf_ablation}(a), and \ref{fig:img_clsf_ablation}(b), we can see that in both cases of m=10 and m=20 respectively the results of CDAL-RL are better than existing techniques.
\setlength{\tabcolsep}{-1pt}
\begin{figure}[h]
	\centering
	\includegraphics[width=0.5\linewidth]{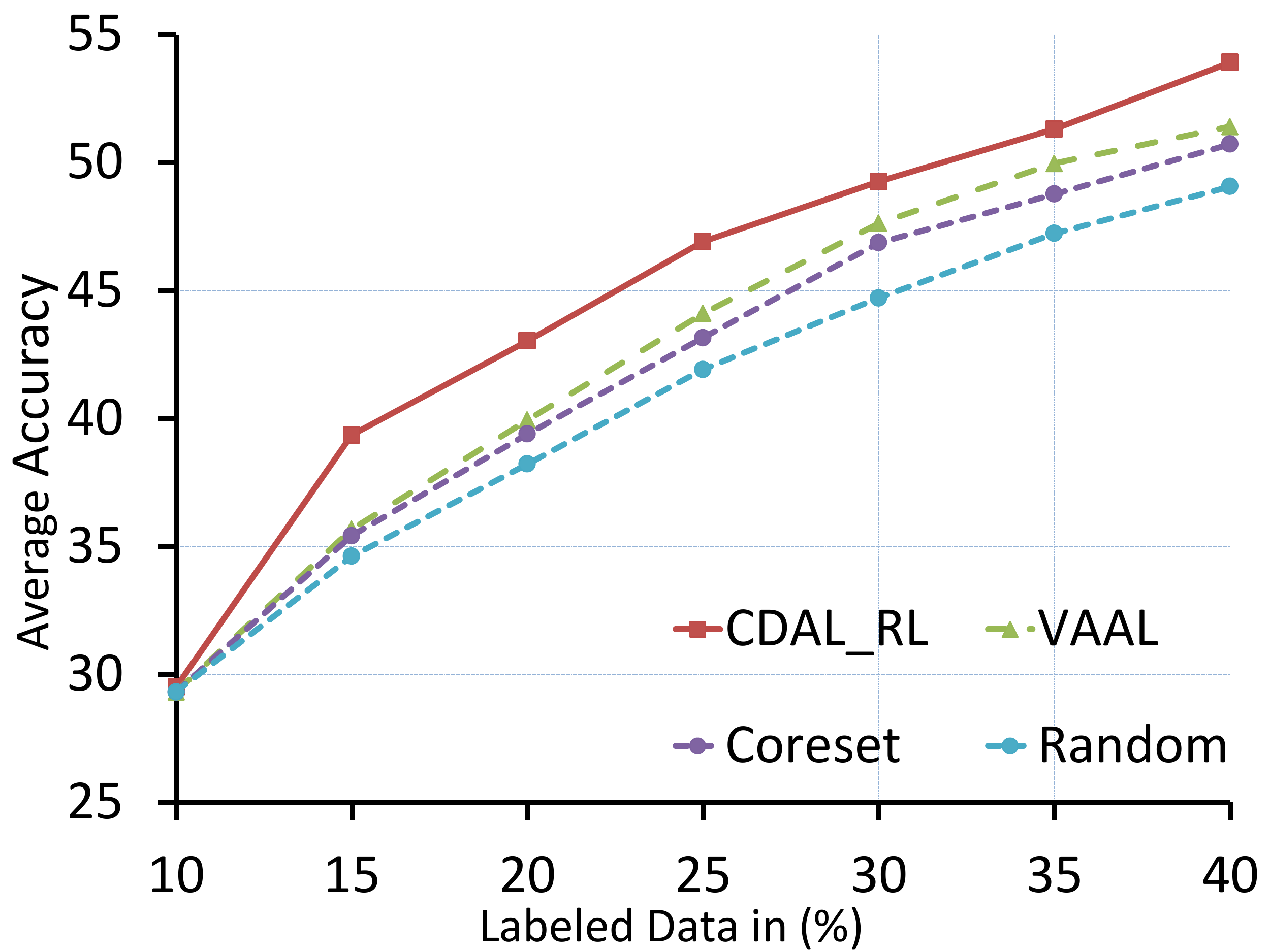}
	\caption{Performance comparison of CDAL-RL using the ResNet-18 architecture on CIFAR-100.}
	\label{fig:resnet18}
\end{figure}

\mypara{Noisy Oracle}
In the second set of experiments, we incorporated noisy oracle, similar to VAAL \cite{VAAL_ICCV2019}. We replaced 10\%, 20\% and 30\% of the selected labels with a random class from the same super-class. Figure \ref{fig:img_clsf_ablation}(c) shows that CDAL-RL is substantially more robust to noisy labels as compared to other approaches. This is possibly due to the pairwise KL-divergence based selection of frames, which effectively captures the disparity of confusion between frames and is less sensitive to outliers.

\mypara{Varying Budget}
In the second set of experiments we changed the step size for varying budget experiments. In the main paper, all the experiments have been shown with a step size of 5\% (e.g. 20\%, 25\%, 30\% and so on). Figure \ref{fig:img_clsf_ablation}(d) shows results on CIFAR 100 with budget steps of 10\%. We can see that varying budget does not much effect the performance of CDAL-RL and performs better than VAAL \cite{VAAL_ICCV2019} and its competitive approaches.

\mypara{Change in network Architecture}
As shown in the case of semantic segmentation CDAL-RL performs better irrespective of the network architecture, here we show results of CIFAR-100 on ResNet18, as shown in Figure \ref{fig:resnet18} our CDAL-RLoutperforms all existing baselines by a substantial margin.

\section{Ablation: Sensitivity analysis of $\alpha$}
As discussed in section 3.2 we have performed all the experiments with $\alpha= 0.75$. Here we justify that the selection of the weight for the reward was not very sensitive, but we focused on giving more weight to the CD component of the reward. The experiment was performed for CDAL-RL using the model trained at 20\%, with the goal of selecting the next 5\% samples to retrain the model at 25\%. As we can see in the Table \ref{tab:ablation_alpha}, we performed experiments with $\alpha$ taking values of 0.25, 0.5 and 0.75 and obtained the highest mIoU for 0.75. Even the smallest mIoU of 55.5\% is still significantly higher than VAAL ($ \sim 54\% $) and CDAL-CS ($ \sim 54.9\% $). Therefore, while the performance may change by increasing $ \alpha $, it still is better than other competing methods. 

\setlength{\tabcolsep}{12pt}
\begin{table}[h!]
	\begin{center}
		\begin{tabular}{cccc}
			\toprule[1.5pt]
			$\alpha$ & 0.25 & 0.50 & 0.75\\
			\toprule[1.5pt]
			mIoU & 55.5 & 55.8 & 56.3\\
			\bottomrule[1.5pt]
		\end{tabular}
	\end{center}
	\caption{Cityscapes ablation of $\alpha$ weighting factor for reward in CDAL-RL}
	\label{tab:ablation_alpha}
\end{table}

\subsection{Weights ablation of mixture distribution}
Following the same experimental setup as above, we used the Cityscapes dataset and the model trained using 20\% of teh data. With the goal of making the selection of the next 5\% samples, we changed the mixture weights in Eq. (1) instead of Shannon's entropy. As we result, there was a deterioration in the 25\% performance from 56.3\% mIoU to 55.2\% mIoU. This is in line with our expectation that the Shannon's entropy used as weight better captures the uncertainty in the predictions and therefore leads to better selections. 

\section{Algorithm}
\begin{algorithm}[H]
	\caption{CDAL-CS}
	\label{alg:CDAL_CS}
	{\textbf{Input: } Unlabelled pool features $X^{u}$, Budget b, selected pool $s$}
	
	\begin{algorithmic}[1]
		\State Add randomly selected data point $x_{0}\in X^u$ to $s$
		\State Initialize a distance matrix $D$ of size $ |S|\times |X^u| $ using Eq.(2) as distance metric.
		\State \textbf{repeat}
		\State $\>$ $\>$ compute $ \widehat{D} $ as a $ |X^u| $ dimensional vector of minimum distances from each centroid
		\State $\>$ $\>$ select new centroid using $ u = \arg\max(\widehat{D}) $ 
		\State $\>$ $\>$ add $u$ to selected pool $s$
		\State $\>$ $\>$ update D 
		\State \textbf{until}   $  |s| = |b| $
		\State \textbf{return}  $s$ 
	\end{algorithmic}
\end{algorithm}

\begin{algorithm}[H]
	\caption{CDAL-RL}
	\label{alg:CDAL_RL}
	{\textbf{Input: } Unlabelled pool features $X^{u}$, RL Model Parameters $\theta_{RL}$}
	\begin{algorithmic}[1]
		
		\State \textbf{for} e = 1 to epochs \textbf{do} 
		\State  $\>$ $\>$ Predict $p_{t}$ for every data point 
		\State  $\>$ $\>$ sample $x^{u}\sim X^{u}$ using highest probabilities
		\State  $\>$ $\>$ compute $R_{cd}$ , $R_{vr}$ , $R_{sr}$
		\State  $\>$ $\>$ Using REINFORCE algorithm, calculate gradient $\nabla_{\theta}J(\theta)$
		\State  $\>$ $\>$ $\nabla_{\theta}J(\theta) = \frac{1}{N} \sum \sum R_{n}$ 
		\State  $\>$ $\>$ Update $\theta_{RL}$ using SGD
		\State  $\>$ $\>$ $\theta_{RL} = \theta_{RL} - \alpha\nabla_{\theta}(-J)$
		\State \textbf{return} trained $\theta_{RL}$
		
	\end{algorithmic}
\end{algorithm}

\end{document}